\documentclass[10pt,twocolumn,letterpaper]{article}

\usepackage{iccv}

\usepackage{times}
\usepackage{epsfig}
\usepackage{amsmath}
\usepackage{amssymb}

\usepackage{booktabs}
\usepackage{xcolor}
\def\R{\mathbb{R}}
\def\G{\mathcal{G}}
\def\V{\mathcal{V}}
\def\E{\mathcal{E}}

\def\P{\mathcal{R}}

\newcommand{\M}{RCRN}
\newcommand{\D}{GITM-MR}
\newcommand{\T}{GITM-MR}

\usepackage{bm}
\usepackage{algorithm}
\usepackage{algpseudocode}
\usepackage{bbm}

\usepackage{graphicx} 
\usepackage{float} 
\usepackage{multirow}
\usepackage{subfigure}
\usepackage{makecell}

\usepackage[pagebackref=true,breaklinks=true,letterpaper=true,colorlinks,bookmarks=false]{hyperref}
\pdfoutput=1  
\iccvfinalcopy 

\def\iccvPaperID{8957} 

\ificcvfinal\fi

\begin{document}

\title{Grounded Image Text Matching with Mismatched Relation Reasoning}

\author{Yu Wu\textsuperscript{\rm1}\footnotemark[1]  \space
	Yana Wei\textsuperscript{\rm1}\footnotemark[1]  \space
	Haozhe Wang\textsuperscript{\rm2} \space
	Yongfei Liu\textsuperscript{\rm3} \space
	Sibei Yang\textsuperscript{\rm1,4} \space
	Xuming He\textsuperscript{\rm1,4}\\
	ShanghaiTech University, Shanghai, China\textsuperscript{\rm1}
	Alibaba Group\textsuperscript{\rm2}  ByteDance Inc.\textsuperscript{\rm3}  \\
	Shanghai Engineering Research Center of Intelligent Vision and Imaging\textsuperscript{\rm4} \\ 
	{\tt\small {\{wuyu1, weiyn1, yangsb, hexm\}@shanghaitech.edu.cn\space jasper.whz@outlook.com\space liuyongfei314@gmail.com}}
\and 
}
\maketitle

\renewcommand{\thefootnote}{\fnsymbol{footnote}}
\footnotetext[1]{Both authors contributed equally to this work, which was supported by Shanghai Science and Technology Program 21010502700, Shanghai Frontiers Science Center of Human-centered Artificial Intelligence and MoE Key Lab of Intelligent Perception and Human-Machine Collaboration (ShanghaiTech University).}

\maketitle
\ificcvfinal\thispagestyle{empty}\fi

\begin{abstract}
	This paper introduces Grounded Image Text Matching with Mismatched Relation (GITM-MR), a novel visual-linguistic joint task that evaluates the relation understanding capabilities of transformer-based pre-trained models. GITM-MR requires a model to first determine if an expression describes an image, then localize referred objects or ground the mismatched parts of the text. We provide a benchmark for evaluating pre-trained models on this task, with a focus on the challenging settings of limited data and out-of-distribution sentence lengths. Our evaluation demonstrates that pre-trained models lack data efficiency and length generalization ability. To address this, we propose the Relation-sensitive Correspondence Reasoning Network (RCRN), which incorporates relation-aware reasoning via bi-directional message propagation guided by language structure. RCRN can be interpreted as a modular program and delivers strong performance in both length generalization and data efficiency. 
\end{abstract}
\section{Introduction}

Recently, transformer-based vision-language (VL) pre-trained models have made significant progress on various VL tasks by fine-tuning on downstream tasks~\cite{chen2020uniter, zhang2021vinvl, yang2022vision, kamath2021mdetr, dou2022coarse}. Despite their success, the representation capacity of such VL pre-trained models remains poorly understood. As a result, an increasing number of studies start to probe the limitations of those learned models from several aspects. 
The first aspect focuses on the lack of fine-grained understanding of multi-modal data, which is indicated by the weak VL correspondence learned on relations among the entities~\cite{nikolaus2022vision}. In particular, the researchers constructed subtle relation phrase differences in text to reveal the poor matching capability between visual relations and linguistic relations~\cite{shekhar2017foil, hendricks2021probing,  parcalabescu2022valse}, or required the model to output fine-grained predictions to thoroughly examine the model's understanding of VL correspondence on different uni-modal component representations~\cite{thomas2022fine}. 
Additionally, while large models generally deliver superior performance, they tend to overfit on small datasets, emphasizing the need to enhance the data-efficiency on fine-tuning~\cite{dodge2020fine}.
Furthermore, when the text is composed of multiple relations, inferring VL correspondence typically involves combining those semantic relations. This combination requires the model to reason the global semantic correspondence, where the length generalization is shown to be a critical weakness in reasoning capability of pre-trained transformers~\cite{anil2022exploring, varivs2021sequence}. However, most of those studies only focus on  a single aspect of the VL models and hence produce a partial view on their limitations.  


\begin{figure}[t]
    \centering
    \includegraphics[width=0.4\textwidth]{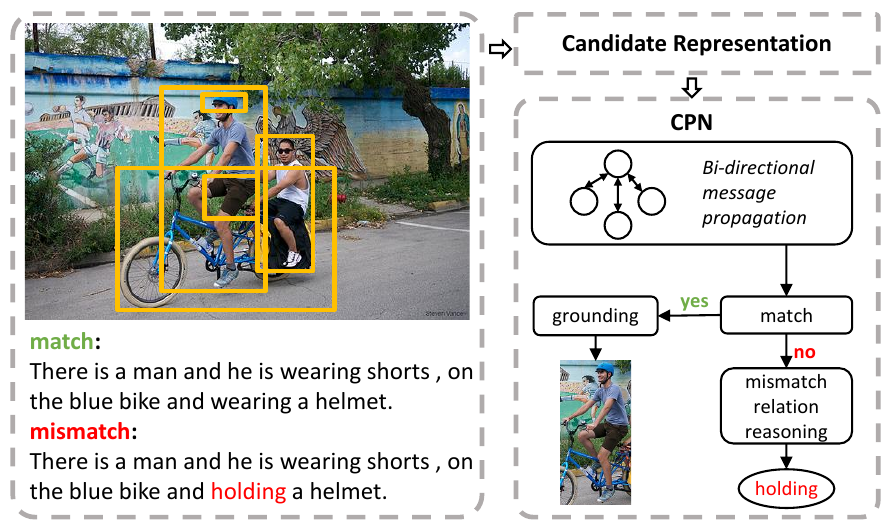}
    \caption{
   Illustration of the \T~task and the proposed model \M. 
   The task is to first identify whether the expression describes the image, and then localize the referred object if the pair are matched, or otherwise ground the mismatched language part. Both data efficiency and length generalization are required.
}
    \label{fig:adv}
    \vspace{-6mm}
\end{figure}

To better understand the VL pre-trained models, we introduce a novel VL joint task named \emph{Grounded Image Text Matching with Mismatched Relation (\T)} which focuses on distinguishing nuance in relations within text and image, and requires a model to show explicit comprehension on relations in both modalities. 
As shown in Fig.~\ref{fig:adv}, the task first requires the model to identify whether an expression describes a given image, and then grounds the referred object if the image-text pair are matched or localizes the mismatched relation in the expression otherwise. The two fine-grained sub-tasks verify the specific understanding on relations both in match and mismatch scenarios. 
Moreover, the proposed task also allows us to evaluate the data efficiency of model's fine-tuning and its generalization ability. In particular, we design a learning setting with limited annotations, where only the data in a limited size and domain have annotations (i.e., a mix of in- and out-of-distribution (OOD) test data). 
Given the task design, we then build a new benchmark based on the Ref-Reasoning~\cite{yang2020graph} dataset, which is more general and challenging than previous probing benchmarks on relations and with low linguistic bias. 

Leveraging our proposed benchmark, we evaluate several representative state-of-the-art models on the task, assessing their performance under limited data sizes and varying domains of input length. Our results reveal their limitations in the both settings. To alleviate these drawbacks, we propose a unified modular framework that jointly solves subtasks in the \T~problem.
By explicitly capturing relation correspondence and utilizing it to propagate messages, our approach is capable of obtaining a comprehensive understanding of global correspondence. Additionally, its lightweight and compositional module design allows for excellent data efficiency and length generalization, accompanied by a more transparent reasoning process.

Specifically, we develop a graph neural network on a parsed language scene graph (LSG), named \emph{Relation-sensitive Correspondence Reasoning Network (\M)}, which exploits the language structure to perform relation-aware reasoning in a belief space of visual-linguistic alignment.
Our \M~uses a set of primitive reasoning modules to operate on the pre-trained vision-language features under the guidance of LSG, and predict the fine-level grounding in both match and mismatch scenarios. 
As demonstrated by the experiments, our method outperforms in both data efficiency and length generalization settings, demonstrating its effectiveness in learning relation correspondences with data efficiency and OOD generalization capability.


The contribution of our work is summarized as follows:
\begin{itemize}
\setlength{\itemsep}{0pt}%
\setlength{\parskip}{0pt}%
\item We propose a new VL task \T~that challenges relation correspondence learning in VL pre-trained model, and focusing on both data efficiency and out-of-distribution text length setting, with a new benchmark for \D.
\item We identify the shortcomings of current VL models through the benchmark, and develop a modular graph network \M~to address these issues.
\item Our method achieves superior results on both settings, validating its relation learning efficiency and generalization ability, which hopefully inspires the following exploration of these challenges. 
\end{itemize}

\section{Related Works}
\vspace{-1mm}
\paragraph{Probing Tasks for VL Models}
Despite the rapid development on VL pre-trained models~\cite{chen2020uniter, radford2021learning, gan2020large, zhang2021vinvl, yang2022vision, kamath2021mdetr}, emergent works reveal the limitations and gain some deep understanding on pre-trained representations, including implausible scenes understanding~\cite{choi2012context}, language and vision priors exploration~\cite{goyal2017making}, and modality ablation~\cite{frank2021vision}.
A considerable number of them have revealed the weakness of VL models in relation understanding.  
Early work proposed a benchmark FOIL ~\cite{shekhar2017foil, shekhar2017vision} by substituting a word in caption to test non-pretrained VL models on identifying minor differences. Subsequent works~\cite{hendricks2021probing, parcalabescu2022valse, nikolaus2022vision, thrush2022winoground} further evaluate pre-trained models on detecting subtle differences in relations or predicate-noun dependencies in zero-shot scenarios, and reach a consensus on the poor quality of pre-trained VL representation for relations. FGVE~\cite{thomas2022fine} extends this evaluation to the task of fine-grained entailment with model fine-tuning, but lacks evaluation in vision modality. Recently, a new benchmark~\cite{salin2022vision} is proposed to simultaneously assess the concept understanding of representations learned from both modalities. In contrast, our task focuses on generating fine-grained multi-modal output under different fine-tuning settings in order to systematically investigate the relation understanding problem in VL pre-training.

\vspace{-4mm}
\paragraph{OOD Generalization}
The general problem of OOD generalization has attracted much attention recently~\cite{li2018deep, buhlmann2020invariance, shen2020disentangled, kuang2020stable, shen2020stable}. 
Among them, several studies have considered the tasks with generalization to processing or generating longer sequences. 
Some works~\cite{anil2022exploring, varivs2021sequence} have studied length generalization in different uni-modal seq-to-seq tasks, and find that fine-tuning pre-trained transformer models can lead to poor out-of-distribution (OOD) performance. 
To address this issue, various solutions have been proposed, including modifications to attention mechanism~\cite{press2021train, dastaircase, dubois2019location} and the design of recurrent networks~\cite{schwarzschild2021can, bansal2022end}, but their application in VL tasks is nontrivial.
Other recent works have also revealed the challenge of OOD generalization for the VL tasks, 
but most of them focuses on the answer bias of Visual Question Answering (VQA)~\cite{teney2020value, agrawal2018don, akula2022question}. The attempted solutions include the causal learning~\cite{niu2021counterfactual} and stable learning~\cite{teney2021unshuffling} methods. Perhaps most related to us is the modular network approaches~\cite{johnson2017inferring,shi2019explainable,zhao2021proto}, which use program-based design to achieve good generalization for certain OOD scenario such as question domains in VQA. We follow the modular design but focus on the OOD problem induced by the expression complexity in the grounded image text matching task. 

\vspace{-4mm}
\paragraph{Vision-language Matching}
Image-Text Matching (ITM) and Referring Expression Grounding (REG) are common cross-modal matching tasks in vision-language domain.
The problem of ITM aims to predict a global similarity between a text and an image.
Classical methods for ITM follow the core idea of learning a common space where linguistic and visual features are compatible~\cite{ji2019saliency,li2019visual,chen2020expressing, diao2021similarity,cheng2022vista}. 
Recent approach~\cite{radford2021learning} concentrates on learning the relation-aware global similarity by a regularization strategy on training. Our model derives a global similarity measure by propagating messages on local correspondences, capable of capturing subtle changes in relations.

The goal of REG is to locate an object in an image based on a language expression.
Most methods focus on learning a holistic visual and linguistic representation in a joint space for aligning the object and text, based on CNN+RNN architectures~\cite{nagaraja2016modeling, zhuang2018parallel, huang2021look} or the Transformer~\cite{deng2021transvg, deng2022transvg++, li2021referring, qu2022siri, yang2022improving}. 
Another line of research aims to exploit the linguistic structure in the text for better context modeling~\cite{Yu_2018_CVPR,liu2020learning,yang2019dynamic,wang2022referring}. In particular,
SGMN~\cite{yang2020graph} parses the text into language components and performs reasoning under the guidance of the linguistic structure. 
Our method introduces a flexible contextual representation and a bi-directional propagation to extend correspondence reasoning beyond the task of REG.

\begin{figure}
	\centering
	\includegraphics[width=0.4\textwidth]{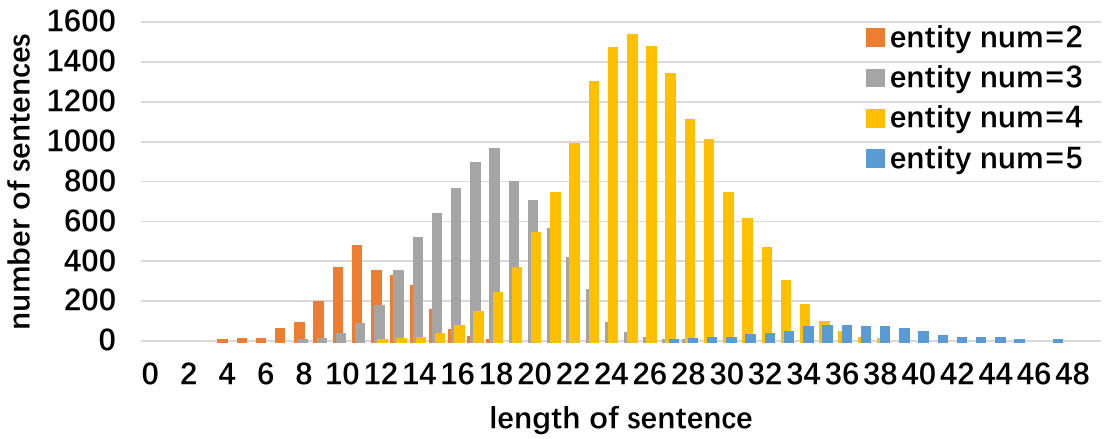}
	\vspace{-1mm}
	\caption{The length distribution of sentences containing different numbers of entities in the validation set. Different colors represent sentences with different number of entities.}
	\vspace{-5mm}
	\label{fig:node and length}
\end{figure}

\begin{figure*}
	\centering
	\includegraphics[width=0.85\textwidth]{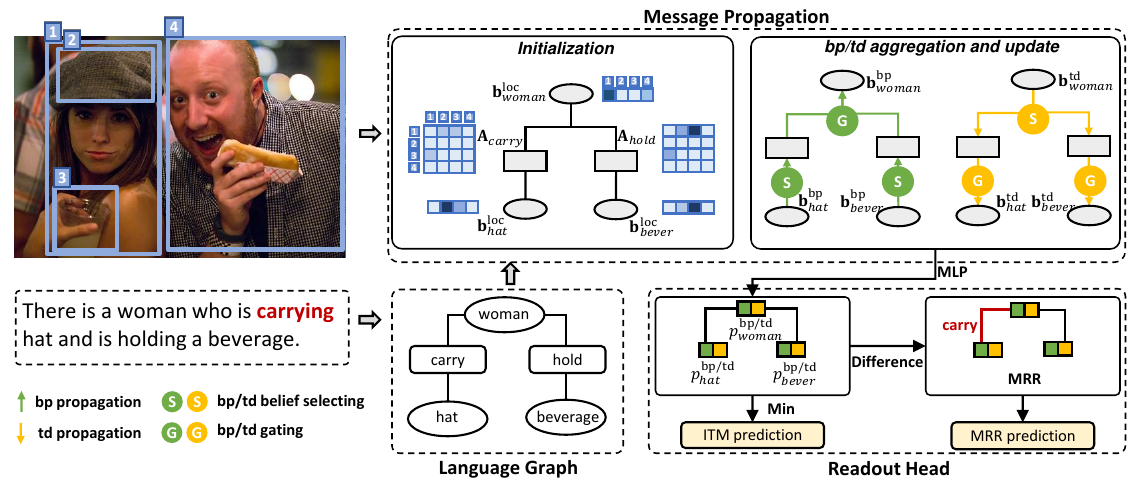}
	\vspace{-1mm}
	
	\caption{\small{Model overview with a mismatched case. Given an image and an expression, we first generate visual and linguistic candidates by a detector and a language parser, and compute their representations. Then we use the Context-sensitive Propagation Network to inter alignments between visual-linguistic candidates, which conducts bi-directional message propagation based on the language graph. The propagation initializes the messages by computing local beliefs, selectively aggregates the context information and updates the belief with a context-sensitive gating function. Predictions for this case are obtained by exploiting the beliefs from the propagation. In the language graph, the ellipses represent entity phrases and the rectangles stand for relation phrases. Further elaboration can be found in the main text.}
	}
	\label{fig:crf}
	\vspace{-5mm}
\end{figure*}

\section{Problem Setup and Benchmark}\label{sec:setup}
We aim to study the representation capacity of the VL pretrained models for fine-grained downstream tasks. To this end, we introduce the \emph{Grounded Image Text Matching with Mismatched Relation (\T)} task, which is a joint matching and grounding task focusing on relations in vision and language domain. In this section, we present the problem setting of \T, followed by building a new benchmark for the task.

\vspace{-4mm}
\paragraph{\T~Task}
Given an image and a referring expression, our goal is to determine if the expression describes an object in the image or not (termed as \emph{Image-Text Matching (ITM)}), and then to localize the referred object if it is matched (termed as \emph{Referring Expression Grounding (REG)}), or otherwise ground the mismatched relation phrase (termed as \emph{Mismatched Relation Reasoning (MRR)}). Notably, the only difference between match and mismatch is a relation phrase in the expression, making our task particularly challenging.

Formally, given an image $I$ and a text description $L$, we denote their matching state as $y$, indicating whether the image matches the description ($y=1$) or not ($y=0$). For the matched case, the referred object has a corresponding location in the image, represented by its bounding box  $z \in \R^4$. 
For the mismatch case, we denote the set of all possible relation phrases (i.e. predicates) for mismatch reasoning as $\P=\left\{{r_i}\right\}_{i=1}^{N_e}$, and there is a target mismatched relation $r_{i}^{\ast}$ in the relation phrase set $\P$. 
Given the input $I$ and $L$, our goal is to first predict ${y}$ for the input pair, and then predict ${z}$ for the referred object in $L$ when $y=1$, or otherwise predict ${r_{i}^{\ast}}$ from $\P$ to explain why $y=0$. 

For model training, we follow the convention of ITM and REG tasks, and assume the annotations for those two subtasks are provided. However, it is usually costly to collect annotations for the subtle MRR task. Consequently, we assume a weakly-supervised learning setting where only the matching state label $y$ is given for mismatched data.

\vspace{-4mm}
\paragraph{Our Benchmark}
{To conduct systematical study, we construct a benchmark for the GITM-MR task, including a new dataset and two challenging evaluation settings.} We first build a dataset of image-text pairs from the public REC dataset, Ref-Reasoning \cite{yang2020graph}, by substituting the relation phrases to generate mismatched referring expressions for images. From the provided relation phrases in the LSG offered by the dataset, we manually select a subset of 27 commonly-occurred relations, and assign some contextual close but semantically different ones for each relation in the subset as their mismatched candidates. We control the linguistic bias by several measures, including keeping the relation distribution and checking by language-only model, the details of which are stated in the Suppl.

However, replacing the relation phrases may introduce falsely mismatched cases.
To ensure the quality of our data, we filter out those false mismatch cases in the test set with the help of annotators.
After filtering, the dataset contains more than 1M referring expressions in 60K images. It has 1M, 50K and 10K expression-referent pairs for training, validation, testing, respectively. We only select the original image text pairs which have generated mismatched pairs to keep the matching label balanced.
Fig.~\ref{fig:node and length} shows the length distribution of sentences in the validation set of the benchmark. {We find it is more general and challenging than previous benchmarks on relation understanding, and refer the reader to see more details of the dataset in Suppl.}

\vspace{-4mm}
\paragraph{Evaluation Settings} 
Given the dataset, we develop two evaluation settings to test pretrained VL models on their relation understanding. 
The first setting is designed to investigate data efficiency during the fine-tuning process. We train the models on small-scale datasets and tested on in-distribution test sets. 
The second setting is designed to evaluate the generalization performance of models with regards to different text input lengths. Here the test set contains longer and more complex samples on relation combinations compared to the training set. 

Specifically, according to the sentence length distribution in Fig.~\ref{fig:node and length}, we construct two training sets, named Train-Len16 and Train-Len11, which includes sentences containing less than 16 and 11 words respectively. In the test phase, the model needs to perform the reasoning task on sentences with lengths less than 11 and 16 (in-distribution), as well as sentences longer than these lengths (out-of-distribution). As two training sets accounts for $5\%$ and $10\%$ of the overall training data, those two settings allows us to evaluate
the model's data efficiency in fine-tuning and OOD generalization ability simultaneously. Detailed data partition rules are explained in Suppl.

\section{Our Approach}\label{sec:model}
In this section, we present our model design for the \T~task and the learning objectives for training with weakly-annotated data. To tackle the problem, we develop a unified and generalizable framework as outlined in Fig.~\ref{fig:adv}, which consists of two main components: a representation network that computes an initial representation of the linguistic and visual components, and a context-aware correspondence reasoning network for final match predictions. Below we first introduce the two main model components in Sec.~\ref{sec:feat} and \ref{sec:CPN}, and then the model learning in Sec.~\ref{sec:learning}.

\subsection{Candidate Generation and Representation} \label{sec:feat}

\subsubsection{Candidate Generation}\label{sec:candidate}
Our method first generates a set of visual objects from the image and phrase candidates from the corresponding text. Specifically, for the visual object candidates, we adopt a pre-trained object detector \cite{zhang2021vinvl} to generate $N_o$ box proposals $ \mathcal{O}=\{o_i\}_{i=1}^{N_o}$ from the image $I$. The detector also provides an initial visual feature (i.e., pooled ROI feature) for each object proposal.
For the language phrase candidates, we utilize an off-the-shelf parser \cite{schuster2015generating} to produce a language scene graph $\G=(\E,\mathcal{R})$ of the text expression $L$, which consists of a set of entity phrases $\E=\{e_i\}_{i=1}^{N_e}$ as its nodes and relation phrases as its edges $\mathcal{R}=\{r_{ij}\}$, where $r_{ij}$ connects to subject $e_i$ and object $e_j$. We denote the number of relation phrases as $|\mathcal{R}|=N_r$. 

\vspace{-4mm}
\subsubsection{Candidate Representation}

Given the visual and linguistic candidates, we compute their representations in two steps, as detailed below.  

\vspace{-4mm}
\paragraph{Token-level Representation}
The first step generates an initial representation for the word tokens and visual objects based on a pre-trained vision-language model. Here we adopt a shallow version of the UNITER model~\cite{chen2020uniter}, aiming to compute a feature representation for the input tokens in two modalities with a generic vision-language alignment from the pre-trained model.\footnote{We note that such a generic representation module has a limited capacity and is typically insufficient for our downstream tasks via a simple fine-tuning procedure as in~\cite{chen2020uniter,yang2022vision}. Nevertheless, it provides rich semantic features for the subsequent alignment reasoning network (c.f. Sec.~\ref{sec:CPN}) after a minimal fine-tuning (described in Sec.~\ref{sec:exp setup}), which complements our explicit reasoning process.}  

\vspace{-4mm}
\paragraph{Candidate Features}
In the second step, we then compile the token-level representations into a set of features for our visual and linguistic components used in the subsequent reasoning network. Here we adopt a similar strategy as in SGMN \cite{yang2020graph} for computing the features.  

Concretely, for visual objects $o_i$, we augment its token-level feature $\mathbf{o}_i$ by its spatial location $\mathbf{l}^o_i$ (i.e., bounding box parameters). In addition, for every object pair ($o_i$, $o_j$) in the image, we represent its relational feature as  $\mathbf{r}_{ij}^{o}$, which is computed from the features of its two objects.
For the linguistic entities in $\E$ and relation phrases in $\mathcal{R}$, we feed the corresponding word tokens into a Bi-LSTM \cite{hochreiter1997long} with self-attention \cite{hu2017modeling} to generate two sets of features. In the first set, each entities $e_i$ is encoded by a tuple of three features $(\mathbf{e}_i, \mathbf{l}^e_i, \mathbf{h}^e_i)$, where $\mathbf{e}_i$ is an embedding into the space of object feature $\mathbf{o}_i$, $\mathbf{l}^e_i$ is an embedding into the space of object location $\mathbf{l}^o_i$, and $\mathbf{h}^e_i$ is a linguistic feature from the LSTM states. For the second feature set, we generate a representation $\mathbf{r}_{ij}^{e}$ for each relation phrase $r_{ij}\in\mathcal{R}$. We refer the reader to Suppl. for the details of our feature design.

\setlength{\abovedisplayskip}{7pt}
\setlength{\belowdisplayskip}{7pt}

\subsection{Context-sensitive Propagation Network}\label{sec:CPN} 

We now introduce our second model component, the Context-sensitive Propagation Network (CPN), which aims to infer the alignment between the visual-linguistic candidates and predict the match label of input image-text pair. 
To this end, we build a modular graph neural network on the language graph $\G$, which takes the candidate features as input and performs context-aware reasoning in a belief space of visual-linguistic alignment. 
Such a structured reasoning process enables us to exploit the regularity in language, i.e. the connections of entities by relations, leading to a relation-aware and compositional alignment model with potentially better generalization.  

Specifically, as illustrated in the language graph in Fig.~\ref{fig:crf}, the $i$-th node of the graph network corresponds to an entity phrase $e_i$ and the edge between the node $i$ and $j$ corresponds to a relation phrase $r_{ij}$. We follow the common assumption that the corresponding $\G$ has a tree structure \cite{yang2020graph, liu2020learning} and the referred entity is regarded as the root node. To perform reasoning, the graph network uses a bi-directional message propagation procedure to infer a relation-aware correspondence belief (w.r.t. the visual candidates) for all the entity nodes (described in Sec.~\ref{sec:program}), followed by a readout head module to predict the fine-grained and global match scores (described in Sec.~\ref{sec:readout}). Finally, we formulate each key operation in the graph reasoning as an easy-to-interpret primitive module, and convert our reasoning process into an explainable program in Sec.~\ref{sec:modular-interpre}.

\vspace{-4mm}
\subsubsection{Bi-directional Message Propagation}
\label{sec:program}

Given the graph network, we aim to compute a message for each node, representing a belief on the correspondence between the associated entity phrase and all the visual objects in the image. In order to  
resolve the ambiguity in the local matching of entities, we introduce a bi-directional message propagation procedure to aggregate all the local beliefs on phrase-visual region correspondence throughout the graph. 

Concretely, inspired by belief propagation~\cite{bishop2006pattern}, we perform two parallel passes of message propagation on the tree graph $\G$ along two directions: 1) bottom-up direction (denoted as $\mathbf{bp}$) first computes messages from the leaf nodes and then updates their parents recursively until the root is reached; 2) top-down direction (denoted as $\mathbf{td}$) starts from the root node and updates the children nodes recursively until reaching all the leaves. 
For each direction, the message propagation aggregates the context information on cross-modal matching and update the belief on phrase alignment at each node in the following three steps:

(1) \textbf{\textit{Message initialization:}} We initialize a local belief $\mathbf{b}^{\rm loc}_i$ at node $i$ based on a similarity measure between the entity $e_i$ and all the visual objects. Here we define the similarity by a weighted combination of feature similarities in appearance and spatial location space. Specifically, for each pair of entity $e_i$ and object $o_j$, we compute two similarities as follows,
\begin{align}
\label{eq:ent-sim}
b_{ij}^{\rm app}=\operatorname{F_{sim}}(\mathbf{e}_i, \mathbf{o}_j), \quad 
b_{ij}^{\rm pos}=\operatorname{F_{sim}}(\mathbf{l}_i^e, \mathbf{l}_j^o) 
\end{align}
where $\operatorname{F_{sim}}$ is an MLP network computing the similarity between features from two modalities. 
The local belief $\mathbf{b}^{\rm loc}_i$ then integrates two similarities as follows:
\begin{align}
\label{eq:local}\mathbf{b}^{\rm loc}_{i}  &= \operatorname{F_{norm}}\left(\beta^{\rm app}_{i} \mathbf{b}_{i}^{\rm app}+\beta^{ \rm pos}_{i} \mathbf{b}_{i}^{\rm pos}\right)
\end{align}
where $\mathbf{b}^{t}_{i}=[b_{ij}^{t}]_{j=1}^{N_o}$, $t \in \{{\rm app},{\rm pos}\}$. Here
$\beta_{i}^{t}$ is the weight for combining similarities and is computed as $\beta_{i}^{t} = \operatorname{sigmoid}\left(\mathbf{W}^{\top}_{t} \left[\mathbf{h}^e_{i}, \mathbf{b}_{i}^{\rm app}, \mathbf{b}_{i}^{\rm pos} \right]\right)$, where $\operatorname{F_{norm}}$ is a normalization function to rescale its input elements to $[0,1]$.  



(2) \textbf{\textit{Message aggregation:}} According to the direction $\mathbf{td}$ or $\mathbf{bp}$, message aggregation collects beliefs from the upstream neighbors in a context-sensitive manner. 
Formally, we first prune the local beliefs by selecting top $K$ maximum values, which significantly reduces the computation complexity for large $N_o$ and is written as,
\begin{align}
\label{eq:selection}
\mathbf{b}^{\rm sel}_j &= \operatorname{F_{topk}}\left(\mathbf{b}^{\rm loc}_{j}\right)
\end{align}
In contrast to \cite{yang2020graph}, our selection is belief-based and more informative for propagation. Given the sparsified beliefs, we then aggregate the messages from the neigbhoring nodes by taking into account the alignment of the relation phrases on the edges as follows,
\begin{align}
\label{eq:fuseprod}
\mathbf{b}_{i}^{\rm agg} = \prod_{j\in\text{Ctx}(i)} \mathbf{A}_{ij} \mathbf{b}^{\rm sel}_j, \quad 
[\mathbf{A}_{ij}]_{kl} = \operatorname{F_{sim}}\left(\mathbf{r}_{ij}^e, \mathbf{r}_{kl}^o\right) 
\end{align}
where $\text{Ctx}(i)$ denotes the upstream neighbor set of node $v_i$ according to the propagation direction. We model the messages provided by relations explicitly through alignment on edges, which forces the model to capture the nuances of different relations.
For the aggregation operator, we take the element-wise product of beliefs, which can be viewed as a soft version of logical AND. Unlike the min pooling used in \cite{shi2019explainable, hu2017learning}, our design leads to a more robust and smooth learning procedure.


(3) \textbf{\textit{Message update:}} Finally, we update the belief $\mathbf{b}_i$ at node $i$ by a context-sensitive integration of the aggregated neighboring belief and the local belief. To achieve this, we design a gated product as follows:
\begin{align}
\label{eq:gateprod}
\beta_{i}^{\rm rel} &= \operatorname{sigmoid}\left(\mathbf{W}^{\top}_{{\rm rel}} \left[\mathbf{h}_{i}^e, \mathbf{b}_{i}^{\rm app}, \mathbf{b}_{i}^{\rm pos} \right]\right)\\
\label{eq:update}
\mathbf{b}_{i} &= \operatorname{F_{norm}}\left(\mathbf{b}^{\rm loc}_{i} \circ (\mathbf{b}_{i} ^{\rm agg})^{\beta_{i}^{\rm rel}}\right) 
\end{align}
where $\circ$ denotes the element-wise product. Instead of using the language features as in \cite{yang2020graph, velickovic2018graph}, our gating function also takes the local beliefs as input, making the network more sensitive to the image context. 

Given the updated beliefs, we compute a confidence score for each node matching to any object in the image by applying an MLP network, denoted as  
	$p_{i} = \operatorname{sigmoid}(\operatorname{MLP_{\rm match}}(\mathbf{b}_{i}))$. 
In the end, the message propagation along two directions generates two  belief vectors ($\mathbf{b}^{\rm bp}_{i}$ and $\mathbf{b}^{\rm td}_{i}$) and two confidence scores ($p^{\text{bp}}_{i}$ and $p^{\text{td}}_{i}$) for each node $i$.


\vspace{-2mm}
\subsubsection{\textbf{Readout Head}}\label{sec:readout}

After the graph-based reasoning, we now introduce a readout head network to generate the predictions for three subtasks, i.e., ITM, REG and MRR, by exploiting the beliefs and confidence scores from the bi-directional propagation.

\vspace{-4mm}
\paragraph{Image-Text Matching} To predict the global matching label $y$, we take a minimum pooling of the matching confidence scores across the entire graph as below, 
\begin{align}
	P(y=1|I,L) =\min_i\min(p^{\text{bp}}_{i},p^{\text{td}}_{i})
\end{align}

\vspace{-5mm}
\paragraph{REG Task} For the referent grounding, we take the belief vector of the root node $\mathbf{b}^{\rm bp}_{\rm root}$ from the bottom-up direction and locate the matching visual object $o_{i^\ast}$ by choosing the largest belief value: ${i^\ast} = \arg\max_{i}{\mathbf{b}^{\rm bp}_{\rm root}}$. A standard regression head is also employed for more precise localization and its details are left to the Suppl. 

\vspace{-4mm}
\paragraph{MRR Task} For mismatched relation prediction, we exploits the inconsistency of the match confidence scores along the graph edges. We found the mismatched relation typically leads to a confidence gap between its two incident nodes. As such, we choose the edge with the largest match confidence difference between its two nodes:
\begin{align}
\label{eq:compare}
\hat{r}=\arg\min_{r_{ij} \in \P}(\{(p^{\rm bp}_{i}-p^{\rm bp}_{j}) + (p^{\rm td}_{j}-p^{\rm td}_{i})\}),
\end{align}
where we sum the differences from two propagation directions to make the decision more robust.

\vspace{-2mm}
\subsubsection{Modular Network Interpretation}\label{sec:modular-interpre}

\begin{table}[t]
	\centering
	\resizebox{0.4\textwidth}{!}{
		\begin{tabular}{|c|c|c|c|}
			\hline
			\textbf{Modules}      & \textbf{In} & \textbf{Out} & \textbf{Implementation} \\ \hline

			\textit{Local Correspondences:}     &     &        &                \\
			\texttt{Sim}                    & $\mathbf{v}^{\rm txt}, \mathbf{v}^{\rm vis}$  &    $b$    &   $\operatorname{F_{sim}}(\mathbf{v}^{\rm txt}, \mathbf{v}^{\rm vis})  $         \\			
			\texttt{GateSum} & $\{\mathbf{v}_i\}, \{
			\mathbf{b}_{i}\} $ & $\mathbf{b}$ & $\operatorname{F_{norm}}\left(\sum_{i} \operatorname{sigmoid}\left(\mathbf{W}^{\top} [\mathbf{v}_i,\mathbf{b}_{i}]\right) 
			\mathbf{b}_{i}\right)$\\ \hline

			\textit{Correspondence Reasoning:} &       &        &                \\
			\texttt{Select} & $\mathbf{b}$ & $\mathbf{b'}$ & $\operatorname{F_{topk}}\left(\mathbf{b}\right)$ \\
			\texttt{Aggregate}                      &  $\{\mathbf{A}_{i}\}, \{\mathbf{b}_{i}\}$     &   $\mathbf{b'}$     &      $\prod_{i}(\mathbf{A}_{i} \mathbf{b}_i)$          \\
			\texttt{GateProd}  & $\mathbf{b},\mathbf{b'}, \beta_{i}$ &  $\mathbf{b''}$ & $\operatorname{F_{norm}}\left(\mathbf{b} \circ (\mathbf{b'})^{\beta_{i}}\right)$ \\
			\texttt{Classify}  & $\mathbf{b}$ &  $p$ & $\operatorname{MLP_{\rm match}}(\mathbf{b})$ \\
			\hline
			
			\textit{Readout Head:}  & & & \\
			\texttt{Locate}  &  $\mathbf{p}$ & $i$ & $ \arg\min(\mathbf{p})$  \\
			\texttt{Compare} &  $p_i, p_j$ & $d$ & $ p_{i}-p_{j} $ \\
			\texttt{And} & $\{p_{i}\}$  & $p$ & $ \min_i p_{i}$ \\ 
			\hline
			
	\end{tabular}}
	\vspace{1mm}
	\caption{\small{The full list of the reasoning modules and their implementations. We use bold lowercase variables to represent vectors and bold uppercase variables to represent matrices. Regular variables are scalars. $\mathbf{W}$ is trainable parameter matrix.}} 
\label{tab:modules}
\vspace{-5mm}
\end{table}

We now introduce an interpretation of our  using the modular network framework. 
As shown in Tab.~\ref{tab:modules}, we summarize the key operations in the \M~into a set of primitive modules with semantic meanings. This allows us to build a explainable reasoning program: (1) In the message initialization step, local beliefs are computed using \texttt{Sim}, and are combined by \texttt{GateSum} to generate the initial belief on each node. (2) For message aggregation, we first use \texttt{Select} to prune each node and then apply \texttt{Aggregate} to collect the messages from all neighbors, considering the alignment of relations particularly. (3) For message update, \texttt{GateProd} is used to combine the aggregated message with the local belief to generate the updated belief on each node. The \texttt{Classify} module provides a confidence score indicating the likelihood that the phrase entity matches an visual object. (4) In the readout head, we use two \texttt{Compare} modules in two directions respectively for MRR and one \texttt{And} operation for the ITM task, and finally generate predictions for all three subtasks via \texttt{Locate} module. For instance, the grounding prediction is achieved by $\texttt{Locate}(-\mathbf{b}_{\rm ref})$. Several example are provided in Suppl. to illustrate the interpretable reasoning process, along with further detailed implementations of the modules.

We note that such a program formulation indicates that the model can enjoy a good out-of-domain generalization due to the compositionality of language graphs. Even facing with more complex text structures, we can perform this reasoning process by reusing those modules on new graphs.

\subsection{Model Learning}\label{sec:learning}
We train the network on a multitask loss with end-to-end back propagation. 
Cross-Entropy (CE) loss and Binary Cross-Entropy (BCE) loss are used for the REG and ITM tasks, respectively:
\begin{equation}
\mathcal{L}_{\rm grd} = \operatorname{CE}\left(\mathbf{p}_{\rm grd}, \mathbf{p}_{\rm grd}^{\ast}\right)
\end{equation}
\begin{equation}
\mathcal{L}_{\rm match} = \operatorname{BCE}\left(p_{\rm match}, y^{\ast}\right)
\end{equation}
where $\mathbf{p}_{\rm grd}=\operatorname{softmax}(\mathbf{b}_{\rm root}^{\rm bp})$ stands for the predicted grounding distribution, $\mathbf{p}_{\rm grd}^{\ast}$ denotes the ground-truth one-hot grounding label, and $p_{\rm match}=P(y=1|I,L)$ denotes the predicted match probability. When the object boxes are from a detector, the box with the maximum IoU with the ground-truth box of the referent is labeled as the localization output.
The overall loss is defined as :
\begin{equation}
    \mathcal{L}=\mathcal{L}_{\rm grd}+\mu \mathcal{L}_{\rm match}+\omega  \mathcal{L}_{\rm reg}
\end{equation}
where $\mu$ and $\omega$ are hyper-parameters to balance the three terms. $\omega$ would be zero if the IoU of the ground-truth box and the selected proposal is less than 0.5.
The regression loss is standard and can be found in Suppl.
\section{Experiments}
\begin{table*}[ht]
	\centering
	\renewcommand\arraystretch{1.3}
	\resizebox{\textwidth}{!}{
		\begin{tabular}{c|c|cc|ccc|ccc|ccc}
			\toprule[1pt]
			\multirow{2}{*}{\textbf{Training Set}}& \multirow{2}{*}{\textbf{Method}} &\textbf{\#Pretrain} & \multirow{2}{*}{\textbf{\#Param}}& \multicolumn{3}{|c|}{\textbf{Full Test}}& \multicolumn{3}{|c|}{\textbf{In-Distribution}}& \multicolumn{3}{|c}{\textbf{Out-of-Distribution}} \\
			 & & \textbf{Images} &    &\textbf{Match\%}&\textbf{Grounding\%}  &\textbf{MRR\%}   &\textbf{Match\%}&\textbf{Grounding\%}  &\textbf{MRR\%}   &\textbf{Match\%}&\textbf{Grounding\%}  &\textbf{MRR\%}\\ 
			
			\hline\hline
			\multirow{6}{*}{Train-Len16}           
            & TCL~\cite{yang2022vision} & 4M & 210M &52.77	&-	&-	&53.11	&-	&-	&52.70	&-	&-\\
			& UNITER~\cite{chen2020uniter} & 4M&111M  &61.93	&24.16	&-	&71.29	&40.63	&-	&59.94	&20.67	&-\\
            & UNITER+MIL~\cite{wang2019comparison} & 4M &111M  &59.13	&18.46	&47.5	&68.83	&41.53	&\textbf{61.38}	&57.07	&13.56	&44.53 \\
            & FIBER~\cite{dou2022coarse} & 4.8M & 254M  &52.31	&9.01   &-	&59.04	&37.15	&-	&50.88	&3.03	&-\\
            & FGVE+MAX~\cite{thomas2022fine} & 5.7M & 113M & 54.30 &  - & 33.23 & 58.37 & - & 51.79 & 53.43 & - & 29.26 \\
            
			& RCRN (Ours)  & 4M & 90M &\textbf{66.59}	&\textbf{29.1}	&\textbf{55.72}	&\textbf{71.85}	&\textbf{46.24}	&59.04	&\textbf{65.47}	&\textbf{25.46}	&\textbf{55.01}\\

			\hline\hline
			\multirow{6}{*}{Train-Len11}            
            & TCL~\cite{yang2022vision} & 4M & 210M &51.71	&-	&-	&52.08	&-	&-	&51.70	&-	&-\\
			& UNITER~\cite{chen2020uniter} & 4M &111M  &55.51	&10.24	&-	&68.22	&30.05	&-	&54.97	&9.42	&-\\
            & UNITER+MIL~\cite{wang2019comparison} & 4M &111M  &55.43	&11.44	&42.73	&65.04	&29.56	&\textbf{61.65}	&55.02	&10.69	&41.93 \\
            & FIBER~\cite{dou2022coarse}& 4.8M &254M &53.52	&11.74 	&-	&57.21	&29.56	&-	&53.36	&11.00	&-\\
            & FGVE+MAX~\cite{thomas2022fine} & 5.7M & 113M &53.40 &- &38.05	&55.26 &- &40.78 &53.31	&- &37.93 \\
			& RCRN (Ours)  & 4M & 90M &\textbf{63.41}	&\textbf{21.45}	&\textbf{52.67}	&\textbf{68.70}	&\textbf{37.93}	&57.77	&\textbf{63.19}	&\textbf{20.77}	&\textbf{52.46}\\
			\bottomrule[1pt]
	\end{tabular}
}
	\vspace{0.1mm}
	\caption{\small{
        Experiment results of testing models' relation understanding with limited data and ability of length generalization.
        UNITER+MIL represents the UNITER with multiple instance learning strategy applied.
        FGVE+MAX represents the fine-grained visual entailment model with max pooling on knowledge elements to predict mismatched relation.
        `-' means the model can't handle the corresponding task.
	}}
	\label{tb:generalize}
	\vspace{-5mm}
\end{table*}
In this section, we conducted experiments on the proposed benchmark to test several baselines of VL pre-trained models on inferring relation correspondences, with a specific focus on their data efficiency and length generalization capabilities. We then compare our model with those strong baselines to show its effectiveness in these settings.

\subsection{Experiment Setup}
\label{sec:exp setup}
\paragraph{Metrics} 
The evaluation metrics include classification accuracy for three subtasks. 
The grounding result for a matched case is considered as correct when it is identified as matching and the predicted box has at least 0.5 IoU with its ground-truth location. The grounding accuracy (i.e. Recall@1) is the ratio of correctly grounded cases. For mismatch reasoning, a mismatched case needs to be correctly classified and the mismatched relation should be accurately selected from the candidate set. The MRR accuracy is the top-1 accuracy among the candidates.


\paragraph{VL Pre-trained Models}
We benchmark five VL models on the proposed dataset. We first choose TCL~\cite{yang2022vision}, UNITER~\cite{chen2020uniter} and FIBER~\cite{dou2022coarse}, where 
TCL is the state-of-the-art pre-trained model for image-text retrieval, UNITER is a vanilla transformer-based VL pre-trained model for matching and grounding tasks, and FIBER uses a specific fine-grained pre-training design. 
Additionally, we adapt the latest fine-grained visual entailment (FGVE) model \cite{thomas2022fine} to our setting. We compare its potential fine-grained multi-task capability with ours, which is designed based on the VL pre-trained model Oscar+\cite{zhang2021vinvl}.  
Because none of the pretrained VL models can fulfill three subtasks jointly, we also apply the multiple instance learning strategy \cite{wang2019comparison} on UNITER to cope with the weakly-supervised MRR task. As shown in Suppl, we modify those models by adding subtask heads to perform three subtasks. 
We note that TCL and UNITER use the same pre-training datasets and the pre-training corpus of FIBER and FGVE is larger while all the models are fine-tuned on the GITM-MR, which leads to a relatively fair comparison. 

\vspace{-4mm}
\paragraph{Implementation Details} 
For generating object candidates, we take the off-the-shelf detector, VinVL \cite{zhang2021vinvl}, to extract at most 100 proposals with object score great than 0.1 for each image. 
For model training, the reasoning component of RCRN is trained by the Adam optimizer with the learning rate set to 5e-4 for the proposed dataset. The learning rate of candidate representation module stated in Sec.~\ref{sec:feat} is set to 5e-7.
We train 80K iterations with batch-size 64 totally. The hyper-parameter $K$, $\mu$ and $\omega$ are set as 5, 3 and 1 based on validation, respectively. The representation module uses the first 6 transformer encoder layers in the pre-trained UNITER. The ablation study on layers and more implementation details are included in the Suppl.
\subsection{Results and Comparisons}
\label{sec:result}

\paragraph{Relation Understanding with Limited Data}

Tab.~\textcolor{red}{\ref{tb:generalize}} (\textbf{In-Distribution} part) shows the performances of pre-trained models on the relation understanding under the condition of limited data.
The results indicate that TCL struggles to achieve a match accuracy of more than chance. This could be attributed to its emphasis on global image-text matching during pre-training, which may hinder its ability to perform fine-grained understanding tasks. 
Though FGVE is designed for fine-grained entailment prediction, it still lacks capability on mismatch relation reasoning in our scenario. This is likely due to its reliance on the pre-trained VL model for the alignment between vision and language relations. 
UNITER and FIBER perform relatively better on the match task than TCL because they both use the World-Region Alignment (WRA) as a pre-training task, and FIBER is also pre-trained with phrase grounding, which helps to establish fine-grained correspondence between image and text. 
The overall conclusion drawn from these findings is that pre-trained models exhibit unsatisfying performance on the challenging benchmark under the limited data condition, indicating the need for further development in establishing effective visual-linguistic correspondence for relation understanding. 

In comparison, our RCRN outperforms the SOTA pre-trained models on match and grounding tasks in the in-distribution scenarios, even with fewer parameters. The model's performance with limited training data indicates the effectiveness of its lightweight and structured design. The design incorporates appropriate inductive biases based on sentence structure to establish both relation and entity correspondence in a visual-linguistic context, and learns the reasoning strategy effectively. 




\vspace{-4mm}
\paragraph{Length Generalization}

Results about length generaliztion ability are shown in the \textbf{Out-of-Distribution} part of Tab.~\textcolor{red}{\ref{tb:generalize}}.  
Overall, the OOD performance of pre-trained models on both test settings was generally poor compared to their in-distribution performance, especially on the grounding and MRR subtasks, which explicitly reflect the models' understanding of relations. 
In particular, 
in the OOD setting, FIBER's match accuracy is similar to TCL, with both models only slightly exceeding 50\%.
The OOD results suggest that these models tend to overfit to the training data biases, without truly establishing the universal VL correspondence of the relations, thus preventing them from generalizing to longer sentences.

In contrast, RCRN achieved notably better performance than the other models on all three subtasks in both OOD scenarios. 
Specifically, RCRN trained with Train-Len11 outperformed the best performance of other models by 8.17\%, 10.08\%, and 10.53\% on all three tasks. 
This success can be attributed to the modular design, which helps the model learn to establish local visual-linguistic correspondence and the universal reasoning strategy. During test, the model can exploit the regularity in sentence structures to compose the modules and deal with longer sentences.

\paragraph{Discussion} 
These experiments not only help us investigate the two challenges that the pre-trained models faced, but also lead to other interesting findings.
First, our observation that fine-grained pre-training tasks, such as WRA, can help models learn local correspondences on downstream tasks, is consistent with the findings of \cite{nikolaus2022vision}. 
Second, our findings in the OOD scenario are consistent with previous works~\cite{anil2022exploring, varivs2021sequence}, which suggest that pre-trained models are prone to overfitting to the training data length biases.

Moreover, to independently investigate the capability on two subtasks, namely grounding and MRR, we also evaluate on an oracle setting, where we suppose that the models always predict correctly on the matching task. The experiment demonstrates our RCRN has strong performance on both subtasks, and the details of this experiment are provided in the Suppl.

Overall, our experiments provide insights into the challenges and potential solutions for the relation understanding in VL models. The fine-grained pre-training tasks can help models learn local correspondences, while the modular design can facilitate the composition of modules to deal with longer sentences. We believe that our findings offer a promising direction to effectively address these challenges.

\subsection{Ablation study}

\begin{table}[t]
    \centering
    \resizebox{0.45\textwidth}{!}{
        \begin{tabular}{cccc|ccc}
            \toprule[1pt]
             \textbf{Lang$\beta$}   &  \textbf{Ctx$\beta$} & \textbf{Bi-MP} & \textbf{VLP} & \textbf{Mat} & \textbf{Grd} & \textbf{MRR}\\ 
            \hline\hline
            -  &   -   &  -  & $\checkmark$(w/o MP)  &54.79     & 13.99   & 41.47 \\ 
            -  &   -   &  $\checkmark$ &  $\checkmark$ &59.65     & 23.80 &	45.58  \\
            $\checkmark$  &  -    &   $\checkmark$ &  $\checkmark$ & 60.88&	24.18 &	49.57 \\
	      
	       $\checkmark$ & $\checkmark$  &  - & $\checkmark$ &60.07 & 23.88 & 48.62 \\
	       $\checkmark$ &  $\checkmark$  &  $\checkmark$  & - & 60.70 &	\textbf{29.25} &	36.65   \\ 
	        \hline
	       $\checkmark$ & $\checkmark$  & $\checkmark$   & $\checkmark$ & \textbf{61.47} &	23.37 &	\textbf{53.23} \\ 
            \bottomrule[1pt]
        \end{tabular}
    }
	\vspace{1mm}
	\caption{\small{Ablation study on the benchmark's OOD validation set.}}
    \vspace{-5mm}
    \label{tb:ablation}
    
\end{table}

In this section, we present the effectiveness of each component in the proposed RCRN through ablative experiments trained on the Train-Len16 set, with the results in Tab.~\textcolor{red}{\ref{tb:ablation}}.

We conduct ablation study by removing some specific designs. \textbf{Lang$\beta$} stands for using the language features as one input for gate functions in Eq.~\ref{eq:local} and Eq.~\ref{eq:update}. Removing it indicates constant $\beta$'s in the propagation. 
\textbf{Ctx$\beta$} means feeding initial similarities into the gate functions.
\textbf{Bi-MP} represents propagating twice individually using different $\beta^{\rm rel}$'s for matching task reasoning. Without this part, only the propagation from the root to leaves is conducted.
\textbf{VLP} stands for introducing vision-language pre-trained transformer layers to encode features. 
Particularly, the first line with VLP only stands for the model which consists of 6 transformer layers from UNITER and the UNITER task heads stated in Sec.~\ref{sec:exp setup} and Sec.~\ref{sec:result}. 

The results show that all of these designs improve the matching performance. 
VLP model (first line in Tab.~\textcolor{red}{\ref{tb:ablation}}) with the same setting as the VLP layers in RCRN performs poorly on OOD scenario, which demonstrates our footnote statement in Sec.~\ref{sec:feat}. Nevertheless, VLP component of RCRN is significant on MRR performance, probably because it provides accurate preliminary vision-language alignments on relations. 

\section{Conclusion}

In this paper, we have introduced a novel VL joint task, Grounded Image Text Matching with Mismatched Relation (GITM-MR), which provides a challenging benchmark for evaluating pre-trained models on the relation reasoning. Our experiments have shown that some state-of-the-art pre-trained models struggle with the task under the setting of limited data and out-of-distribution sentence lengths. To address this problem, we develop a \emph{Relation-sensitive Correspondence Reasoning Network (RCRN)} to compute contextualized cross-modal alignment of both entities and relations, and ground the fine-grained result in the image or expression. The proposed method outperforms prior SOTA pre-trained models, which demonstrates its strong generalization and data efficiency.
Our work sheds light on the limitations of current pre-trained models and the importance of establishing fine-grained vision-language correspondence for relation understanding. 


{\small
\bibliographystyle{ieee_fullname}
\bibliography{egbib}

\begin{thebibliography}{10}\itemsep=-1pt

\bibitem{agrawal2018don}
Aishwarya Agrawal, Dhruv Batra, Devi Parikh, and Aniruddha Kembhavi.
\newblock Don't just assume; look and answer: Overcoming priors for visual
  question answering.
\newblock In {\em Proceedings of the IEEE conference on computer vision and
  pattern recognition}, pages 4971--4980, 2018.

\bibitem{akula2022question}
Arjun~R Akula.
\newblock Question generation for evaluating cross-dataset shifts in
  multi-modal grounding.
\newblock {\em arXiv preprint arXiv:2201.09639}, 2022.

\bibitem{anil2022exploring}
Cem Anil, Yuhuai Wu, Anders Andreassen, Aitor Lewkowycz, Vedant Misra, Vinay
  Ramasesh, Ambrose Slone, Guy Gur-Ari, Ethan Dyer, and Behnam Neyshabur.
\newblock Exploring length generalization in large language models.
\newblock {\em arXiv preprint arXiv:2207.04901}, 2022.

\bibitem{bansal2022end}
Arpit Bansal, Avi Schwarzschild, Eitan Borgnia, Zeyad Emam, Furong Huang, Micah
  Goldblum, and Tom Goldstein.
\newblock End-to-end algorithm synthesis with recurrent networks: Logical
  extrapolation without overthinking.
\newblock {\em arXiv preprint arXiv:2202.05826}, 2022.

\bibitem{bishop2006pattern}
Christopher~M Bishop and Nasser~M Nasrabadi.
\newblock {\em Pattern recognition and machine learning}, volume~4.
\newblock Springer, 2006.

\bibitem{buhlmann2020invariance}
Peter B{\"u}hlmann.
\newblock Invariance, causality and robustness.
\newblock {\em Statistical Science}, 35(3):404--426, 2020.

\bibitem{chen2020expressing}
Tianlang Chen and Jiebo Luo.
\newblock Expressing objects just like words: Recurrent visual embedding for
  image-text matching.
\newblock In {\em Proceedings of the AAAI Conference on Artificial
  Intelligence}, volume~34, pages 10583--10590, 2020.

\bibitem{chen2020uniter}
Yen-Chun Chen, Linjie Li, Licheng Yu, Ahmed El~Kholy, Faisal Ahmed, Zhe Gan, Yu
  Cheng, and Jingjing Liu.
\newblock Uniter: Universal image-text representation learning.
\newblock In {\em European conference on computer vision}, pages 104--120.
  Springer, 2020.

\bibitem{cheng2022vista}
Mengjun Cheng, Yipeng Sun, Longchao Wang, Xiongwei Zhu, Kun Yao, Jie Chen,
  Guoli Song, Junyu Han, Jingtuo Liu, Errui Ding, et~al.
\newblock Vista: vision and scene text aggregation for cross-modal retrieval.
\newblock In {\em Proceedings of the IEEE/CVF Conference on Computer Vision and
  Pattern Recognition}, pages 5184--5193, 2022.

\bibitem{choi2012context}
Myung~Jin Choi, Antonio Torralba, and Alan~S Willsky.
\newblock Context models and out-of-context objects.
\newblock {\em Pattern Recognition Letters}, 33(7):853--862, 2012.

\bibitem{dastaircase}
JU Da, Stephen Roller, Sainbayar Sukhbaatar, and Jason~E Weston.
\newblock Staircase attention for recurrent processing of sequences.
\newblock In {\em Advances in Neural Information Processing Systems}.

\bibitem{deng2021transvg}
Jiajun Deng, Zhengyuan Yang, Tianlang Chen, Wengang Zhou, and Houqiang Li.
\newblock Transvg: End-to-end visual grounding with transformers.
\newblock In {\em Proceedings of the IEEE/CVF International Conference on
  Computer Vision}, pages 1769--1779, 2021.

\bibitem{deng2022transvg++}
Jiajun Deng, Zhengyuan Yang, Daqing Liu, Tianlang Chen, Wengang Zhou, Yanyong
  Zhang, Houqiang Li, and Wanli Ouyang.
\newblock Transvg++: End-to-end visual grounding with language conditioned
  vision transformer.
\newblock {\em arXiv preprint arXiv:2206.06619}, 2022.

\bibitem{diao2021similarity}
Haiwen Diao, Ying Zhang, Lin Ma, and Huchuan Lu.
\newblock Similarity reasoning and filtration for image-text matching.
\newblock In {\em Proceedings of the AAAI Conference on Artificial
  Intelligence}, volume~35, pages 1218--1226, 2021.

\bibitem{dodge2020fine}
Jesse Dodge, Gabriel Ilharco, Roy Schwartz, Ali Farhadi, Hannaneh Hajishirzi,
  and Noah Smith.
\newblock Fine-tuning pretrained language models: Weight initializations, data
  orders, and early stopping.
\newblock {\em arXiv preprint arXiv:2002.06305}, 2020.

\bibitem{dou2022coarse}
Zi-Yi Dou, Aishwarya Kamath, Zhe Gan, Pengchuan Zhang, Jianfeng Wang, Linjie
  Li, Zicheng Liu, Ce Liu, Yann LeCun, Nanyun Peng, et~al.
\newblock Coarse-to-fine vision-language pre-training with fusion in the
  backbone.
\newblock {\em arXiv preprint arXiv:2206.07643}, 2022.

\bibitem{dubois2019location}
Yann Dubois, Gautier Dagan, Dieuwke Hupkes, and Elia Bruni.
\newblock Location attention for extrapolation to longer sequences.
\newblock {\em arXiv preprint arXiv:1911.03872}, 2019.

\bibitem{frank2021vision}
Stella Frank, Emanuele Bugliarello, and Desmond Elliott.
\newblock Vision-and-language or vision-for-language.
\newblock {\em On Cross-Modal Influence in Multimodal Transformers.(2021). DOI:
  https://doi. org/10.18653/v1/2021. emnlp-main}, 775, 2021.

\bibitem{gan2020large}
Zhe Gan, Yen-Chun Chen, Linjie Li, Chen Zhu, Yu Cheng, and Jingjing Liu.
\newblock Large-scale adversarial training for vision-and-language
  representation learning.
\newblock {\em Advances in Neural Information Processing Systems},
  33:6616--6628, 2020.

\bibitem{goyal2017making}
Yash Goyal, Tejas Khot, Douglas Summers-Stay, Dhruv Batra, and Devi Parikh.
\newblock Making the v in vqa matter: Elevating the role of image understanding
  in visual question answering.
\newblock In {\em Proceedings of the IEEE conference on computer vision and
  pattern recognition}, pages 6904--6913, 2017.

\bibitem{hendricks2021probing}
Lisa~Anne Hendricks and Aida Nematzadeh.
\newblock Probing image-language transformers for verb understanding.
\newblock In {\em Findings of the Association for Computational Linguistics:
  ACL-IJCNLP 2021}, pages 3635--3644, 2021.

\bibitem{hochreiter1997long}
Sepp Hochreiter and J{\"u}rgen Schmidhuber.
\newblock Long short-term memory.
\newblock {\em Neural computation}, 9(8):1735--1780, 1997.

\bibitem{hu2017learning}
Ronghang Hu, Jacob Andreas, Marcus Rohrbach, Trevor Darrell, and Kate Saenko.
\newblock Learning to reason: End-to-end module networks for visual question
  answering.
\newblock In {\em Proceedings of the IEEE international conference on computer
  vision}, pages 804--813, 2017.

\bibitem{hu2017modeling}
Ronghang Hu, Marcus Rohrbach, Jacob Andreas, Trevor Darrell, and Kate Saenko.
\newblock Modeling relationships in referential expressions with compositional
  modular networks.
\newblock In {\em Proceedings of the IEEE conference on computer vision and
  pattern recognition}, pages 1115--1124, 2017.

\bibitem{huang2021look}
Binbin Huang, Dongze Lian, Weixin Luo, and Shenghua Gao.
\newblock Look before you leap: Learning landmark features for one-stage visual
  grounding.
\newblock In {\em Proceedings of the IEEE/CVF Conference on Computer Vision and
  Pattern Recognition}, pages 16888--16897, 2021.

\bibitem{hudson2019gqa}
Drew~A Hudson and Christopher~D Manning.
\newblock Gqa: A new dataset for real-world visual reasoning and compositional
  question answering.
\newblock In {\em Proceedings of the IEEE/CVF conference on computer vision and
  pattern recognition}, pages 6700--6709, 2019.

\bibitem{ji2019saliency}
Zhong Ji, Haoran Wang, Jungong Han, and Yanwei Pang.
\newblock Saliency-guided attention network for image-sentence matching.
\newblock In {\em Proceedings of the IEEE/CVF international conference on
  computer vision}, pages 5754--5763, 2019.

\bibitem{johnson2017inferring}
Justin Johnson, Bharath Hariharan, Laurens Van Der~Maaten, Judy Hoffman, Li
  Fei-Fei, C Lawrence~Zitnick, and Ross Girshick.
\newblock Inferring and executing programs for visual reasoning.
\newblock In {\em Proceedings of the IEEE international conference on computer
  vision}, pages 2989--2998, 2017.

\bibitem{kamath2021mdetr}
Aishwarya Kamath, Mannat Singh, Yann LeCun, Gabriel Synnaeve, Ishan Misra, and
  Nicolas Carion.
\newblock Mdetr-modulated detection for end-to-end multi-modal understanding.
\newblock In {\em Proceedings of the IEEE/CVF International Conference on
  Computer Vision}, pages 1780--1790, 2021.

\bibitem{kenton2019bert}
Jacob Devlin Ming-Wei~Chang Kenton and Lee~Kristina Toutanova.
\newblock Bert: Pre-training of deep bidirectional transformers for language
  understanding.
\newblock In {\em Proceedings of NAACL-HLT}, pages 4171--4186, 2019.

\bibitem{kuang2020stable}
Kun Kuang, Ruoxuan Xiong, Peng Cui, Susan Athey, and Bo Li.
\newblock Stable prediction with model misspecification and agnostic
  distribution shift.
\newblock In {\em Proceedings of the AAAI Conference on Artificial
  Intelligence}, volume~34, pages 4485--4492, 2020.

\bibitem{li2019visual}
Kunpeng Li, Yulun Zhang, Kai Li, Yuanyuan Li, and Yun Fu.
\newblock Visual semantic reasoning for image-text matching.
\newblock In {\em Proceedings of the IEEE/CVF International conference on
  computer vision}, pages 4654--4662, 2019.

\bibitem{li2021referring}
Muchen Li and Leonid Sigal.
\newblock Referring transformer: A one-step approach to multi-task visual
  grounding.
\newblock {\em Advances in Neural Information Processing Systems},
  34:19652--19664, 2021.

\bibitem{li2018deep}
Ya Li, Xinmei Tian, Mingming Gong, Yajing Liu, Tongliang Liu, Kun Zhang, and
  Dacheng Tao.
\newblock Deep domain generalization via conditional invariant adversarial
  networks.
\newblock In {\em Proceedings of the European Conference on Computer Vision
  (ECCV)}, pages 624--639, 2018.

\bibitem{liu2020learning}
Yongfei Liu, Bo Wan, Xiaodan Zhu, and Xuming He.
\newblock Learning cross-modal context graph for visual grounding.
\newblock In {\em Proceedings of the AAAI Conference on Artificial
  Intelligence}, volume~34, pages 11645--11652, 2020.

\bibitem{mao2016generation}
Junhua Mao, Jonathan Huang, Alexander Toshev, Oana Camburu, Alan~L Yuille, and
  Kevin Murphy.
\newblock Generation and comprehension of unambiguous object descriptions.
\newblock In {\em Proceedings of the IEEE conference on computer vision and
  pattern recognition}, pages 11--20, 2016.

\bibitem{nagaraja2016modeling}
Varun~K Nagaraja, Vlad~I Morariu, and Larry~S Davis.
\newblock Modeling context between objects for referring expression
  understanding.
\newblock In {\em European Conference on Computer Vision}, pages 792--807.
  Springer, 2016.

\bibitem{nikolaus2022vision}
Mitja Nikolaus, Emmanuelle Salin, Stephane Ayache, Abdellah Fourtassi, and
  Benoit Favre.
\newblock Do vision-and-language transformers learn grounded predicate-noun
  dependencies?
\newblock {\em arXiv preprint arXiv:2210.12079}, 2022.

\bibitem{niu2021counterfactual}
Yulei Niu, Kaihua Tang, Hanwang Zhang, Zhiwu Lu, Xian-Sheng Hua, and Ji-Rong
  Wen.
\newblock Counterfactual vqa: A cause-effect look at language bias.
\newblock In {\em Proceedings of the IEEE/CVF Conference on Computer Vision and
  Pattern Recognition}, pages 12700--12710, 2021.

\bibitem{parcalabescu2022valse}
Letitia Parcalabescu, Michele Cafagna, Lilitta Muradjan, Anette Frank, Iacer
  Calixto, and Albert Gatt.
\newblock Valse: A task-independent benchmark for vision and language models
  centered on linguistic phenomena.
\newblock In {\em Proceedings of the 60th Annual Meeting of the Association for
  Computational Linguistics (Volume 1: Long Papers)}, pages 8253--8280, 2022.

\bibitem{press2021train}
Ofir Press, Noah~A Smith, and Mike Lewis.
\newblock Train short, test long: Attention with linear biases enables input
  length extrapolation.
\newblock {\em arXiv preprint arXiv:2108.12409}, 2021.

\bibitem{qu2022siri}
Mengxue Qu, Yu Wu, Wu Liu, Qiqi Gong, Xiaodan Liang, Olga Russakovsky, Yao
  Zhao, and Yunchao Wei.
\newblock Siri: A simple selective retraining mechanism for transformer-based
  visual grounding.
\newblock In {\em European Conference on Computer Vision}, pages 546--562.
  Springer, 2022.

\bibitem{radford2021learning}
Alec Radford, Jong~Wook Kim, Chris Hallacy, Aditya Ramesh, Gabriel Goh,
  Sandhini Agarwal, Girish Sastry, Amanda Askell, Pamela Mishkin, Jack Clark,
  et~al.
\newblock Learning transferable visual models from natural language
  supervision.
\newblock In {\em International Conference on Machine Learning}, pages
  8748--8763. PMLR, 2021.

\bibitem{salin2022vision}
Emmanuelle Salin, Badreddine Farah, St{\'e}phane Ayache, and Benoit Favre.
\newblock Are vision-language transformers learning multimodal representations?
  a probing perspective.
\newblock In {\em Proceedings of the AAAI Conference on Artificial
  Intelligence}, volume~36, pages 11248--11257, 2022.

\bibitem{schuster2015generating}
Sebastian Schuster, Ranjay Krishna, Angel Chang, Li Fei-Fei, and Christopher~D
  Manning.
\newblock Generating semantically precise scene graphs from textual
  descriptions for improved image retrieval.
\newblock In {\em Proceedings of the fourth workshop on vision and language},
  pages 70--80, 2015.

\bibitem{schwarzschild2021can}
Avi Schwarzschild, Eitan Borgnia, Arjun Gupta, Furong Huang, Uzi Vishkin, Micah
  Goldblum, and Tom Goldstein.
\newblock Can you learn an algorithm? generalizing from easy to hard problems
  with recurrent networks.
\newblock {\em Advances in Neural Information Processing Systems},
  34:6695--6706, 2021.

\bibitem{shekhar2017vision}
Ravi Shekhar, Sandro Pezzelle, Aur{\'e}lie Herbelot, Moin Nabi, Enver
  Sangineto, and Raffaella Bernardi.
\newblock Vision and language integration: Moving beyond objects.
\newblock In {\em IWCS 2017—12th International Conference on Computational
  Semantics—Short papers}, 2017.

\bibitem{shekhar2017foil}
Ravi Shekhar, Sandro Pezzelle, Yauhen Klimovich, Aur{\'e}lie Herbelot, Moin
  Nabi, Enver Sangineto, and Raffaella Bernardi.
\newblock Foil it! find one mismatch between image and language caption.
\newblock In {\em Proceedings of the 55th Annual Meeting of the Association for
  Computational Linguistics (Volume 1: Long Papers)}, pages 255--265, 2017.

\bibitem{shen2020disentangled}
Xinwei Shen, Furui Liu, Hanze Dong, Qing Lian, Zhitang Chen, and Tong Zhang.
\newblock Disentangled generative causal representation learning.
\newblock {\em arXiv preprint arXiv:2010.02637}, 2020.

\bibitem{shen2020stable}
Zheyan Shen, Peng Cui, Tong Zhang, and Kun Kunag.
\newblock Stable learning via sample reweighting.
\newblock In {\em Proceedings of the AAAI Conference on Artificial
  Intelligence}, volume~34, pages 5692--5699, 2020.

\bibitem{shi2019explainable}
Jiaxin Shi, Hanwang Zhang, and Juanzi Li.
\newblock Explainable and explicit visual reasoning over scene graphs.
\newblock In {\em Proceedings of the IEEE/CVF Conference on Computer Vision and
  Pattern Recognition}, pages 8376--8384, 2019.

\bibitem{teney2020value}
Damien Teney, Ehsan Abbasnejad, Kushal Kafle, Robik Shrestha, Christopher
  Kanan, and Anton Van Den~Hengel.
\newblock On the value of out-of-distribution testing: An example of goodhart's
  law.
\newblock {\em Advances in Neural Information Processing Systems}, 33:407--417,
  2020.

\bibitem{teney2021unshuffling}
Damien Teney, Ehsan Abbasnejad, and Anton van~den Hengel.
\newblock Unshuffling data for improved generalization in visual question
  answering.
\newblock In {\em Proceedings of the IEEE/CVF International Conference on
  Computer Vision}, pages 1417--1427, 2021.

\bibitem{thomas2022fine}
Christopher Thomas, Yipeng Zhang, and Shih-Fu Chang.
\newblock Fine-grained visual entailment.
\newblock {\em arXiv preprint arXiv:2203.15704}, 2022.

\bibitem{thrush2022winoground}
Tristan Thrush, Ryan Jiang, Max Bartolo, Amanpreet Singh, Adina Williams, Douwe
  Kiela, and Candace Ross.
\newblock Winoground: Probing vision and language models for visio-linguistic
  compositionality.
\newblock In {\em Proceedings of the IEEE/CVF Conference on Computer Vision and
  Pattern Recognition}, pages 5238--5248, 2022.

\bibitem{varivs2021sequence}
Du{\v{s}}an Vari{\v{s}} and Ond{\v{r}}ej Bojar.
\newblock Sequence length is a domain: Length-based overfitting in transformer
  models.
\newblock {\em arXiv preprint arXiv:2109.07276}, 2021.

\bibitem{velickovic2018graph}
Petar Veli{\v{c}}kovi{\'{c}}, Guillem Cucurull, Arantxa Casanova, Adriana
  Romero, Pietro Li{\`{o}}, and Yoshua Bengio.
\newblock {Graph Attention Networks}.
\newblock {\em International Conference on Learning Representations}, 2018.
\newblock accepted as poster.

\bibitem{wang2022referring}
Jia Wang, Jingcheng Ke, Hong-Han Shuai, Yung-Hui Li, and Wen-Huang Cheng.
\newblock Referring expression comprehension via enhanced cross-modal graph
  attention networks.
\newblock {\em ACM Journal of the ACM (JACM)}, 2022.

\bibitem{wang2019comparison}
Yun Wang, Juncheng Li, and Florian Metze.
\newblock A comparison of five multiple instance learning pooling functions for
  sound event detection with weak labeling.
\newblock In {\em ICASSP 2019-2019 IEEE International Conference on Acoustics,
  Speech and Signal Processing (ICASSP)}, pages 31--35. IEEE, 2019.

\bibitem{yang2022vision}
Jinyu Yang, Jiali Duan, Son Tran, Yi Xu, Sampath Chanda, Liqun Chen, Belinda
  Zeng, Trishul Chilimbi, and Junzhou Huang.
\newblock Vision-language pre-training with triple contrastive learning.
\newblock 2022.

\bibitem{yang2022improving}
Li Yang, Yan Xu, Chunfeng Yuan, Wei Liu, Bing Li, and Weiming Hu.
\newblock Improving visual grounding with visual-linguistic verification and
  iterative reasoning.
\newblock In {\em Proceedings of the IEEE/CVF Conference on Computer Vision and
  Pattern Recognition}, pages 9499--9508, 2022.

\bibitem{yang2019dynamic}
Sibei Yang, Guanbin Li, and Yizhou Yu.
\newblock Dynamic graph attention for referring expression comprehension.
\newblock In {\em Proceedings of the IEEE/CVF International Conference on
  Computer Vision}, pages 4644--4653, 2019.

\bibitem{yang2020graph}
Sibei Yang, Guanbin Li, and Yizhou Yu.
\newblock Graph-structured referring expression reasoning in the wild.
\newblock In {\em Proceedings of the IEEE/CVF Conference on Computer Vision and
  Pattern Recognition}, pages 9952--9961, 2020.

\bibitem{Yu_2018_CVPR}
Licheng Yu, Zhe Lin, Xiaohui Shen, Jimei Yang, Xin Lu, Mohit Bansal, and
  Tamara~L. Berg.
\newblock Mattnet: Modular attention network for referring expression
  comprehension.
\newblock In {\em Proceedings of the IEEE Conference on Computer Vision and
  Pattern Recognition (CVPR)}, June 2018.

\bibitem{zhang2021vinvl}
Pengchuan Zhang, Xiujun Li, Xiaowei Hu, Jianwei Yang, Lei Zhang, Lijuan Wang,
  Yejin Choi, and Jianfeng Gao.
\newblock Vinvl: Revisiting visual representations in vision-language models.
\newblock In {\em Proceedings of the IEEE/CVF Conference on Computer Vision and
  Pattern Recognition}, pages 5579--5588, 2021.

\bibitem{zhao2021proto}
Zelin Zhao, Karan Samel, Binghong Chen, et~al.
\newblock Proto: Program-guided transformer for program-guided tasks.
\newblock {\em Advances in Neural Information Processing Systems},
  34:17021--17036, 2021.

\bibitem{zhuang2018parallel}
Bohan Zhuang, Qi Wu, Chunhua Shen, Ian Reid, and Anton Van Den~Hengel.
\newblock Parallel attention: A unified framework for visual object discovery
  through dialogs and queries.
\newblock In {\em Proceedings of the IEEE Conference on Computer Vision and
  Pattern Recognition}, pages 4252--4261, 2018.

\end{thebibliography}
}
\newpage
\appendix



\section{RCRN} \label{sec:crf}

In this part, we supplement the explanation for the design of features and propagation process in Sec.~\ref{sec:model}. Below we first introduce details about our candidate representations and then further explain functions used in the propagation process.

\subsection{\textbf{Candidate Representation}}
We firstly generate an initial representation for the word tokens and visual objects based on a pre-trained vision-language model, then encode them into candidate representations. Specifically, we use the image embedder in UNITER to encode the convolutional and location features of the object regions in $\mathcal{O}=\left\{o_i \right\}_{i=1}^{N_o}$, and the text embedder to generate an embedding of the word tokens $\mathcal{W}=\left\{w_i\right\}_{i=1}^{N_w}$ and their positions in $L$. Those region and word features are then fed into the first $k$ layers ($k=6$) of Transformer in UNITER to compute their initial cross-modal representations, which are denoted as $\left\{\mathbf{o}_i\right\}_{i=1}^{N_o}$ and 
$\left\{\mathbf{w}_i\right\}_{i=1}^{N_w}$, respectively.

Given the word representations $\left\{\mathbf{w}_i\right\}_{i=1}^{N_w}$ and box features $\left\{\mathbf{o}_i\right\}_{i=1}^{N_o}$ generated by the transformer layers, we compute features for objects and their relations in the visual modality, and phrase features in the language modality as follows.

\paragraph{Visual features.}
For the vision modality, we encode each object by the features of its bounding box region, and each relation by the features in the union box of two objects. The visual object feature for each $o_i$ is the corresponding box region feature $\left\{\mathbf{o}_i\right\}_{i=1}^{N_o}$. Its spatial location feature $\mathbf{l}_{i}^{o}$ is defined as $\left[x_{i}, y_{i}, w_{i}, h_{i}, w_{i} h_{i}\right]$, where $(x_i, y_i)$, $w_{i}$ and $h_{i}$ are the normalized top-left coordinates, width and height of the bounding box of each object $o_i$ respectively.
The visual relation feature $\mathbf{r}_{i j}^{o}$ represents the direct relation between the box pair ($o_i$, $o_j$), which is computed as follows:
\begin{equation}
	\mathbf{r}_{i j}^{o}=\left[\mathbf{W}_{l}^{T} \mathbf{s}_{i j}, \mathbf{W}_{e}^{T}\left[\mathbf{o}_{i},\mathbf{o}_{j}\right]\right], 
\end{equation}
where the relative spatial feature is represented as \\ $\mathbf{s}_{i j}=\left[\frac{x_{j}-x_{c_i}}{w_{i}}, \frac{y_{j}-y_{c_{i}}}{h_{i}}, \frac{x_{j}+w_{j}-x_{c_{i}}}{w_{i}}, \frac{y_{j}+h_{j}-y_{c_{i}}}{h_{i}}, \frac{w_{j} h_{j}}{w_{i} h_{i}}\right]$.  $\mathbf{W}_{l}$ and $\mathbf{W}_{e}$ are both trainable parameters, and ($x_{c_{i}}$, $y_{c_{i}}$) represents the normalized box centers for $o_i$.

\paragraph{Language features.}
The linguistic feature encoding is adapted from the prior work~\cite{yang2020graph}. Guided by the language scene graph, both entity phrases $\E$ and relation phrases $\mathcal{R}$ are encoded by a LSTM and self-attention modules. In particular, the linguistic features for each entity $e_i$ is encoded from words contained by the entity phrase itself, but the representation for each linguistic relation $r_{ij} \in \mathcal{R}$ is computed based on its corresponding subject-predicate-object (SPO) phrase, i.e. ($e_i$, $r_{ij}$, $e_j$).
For a phrase containing $N_p$ words, the initial word representations are $\left\{\mathbf{w}_i\right\}_{i=1}^{N_p}$. First, the initial word representations are fed into a LSTM, and we get hidden vectors of the words $\left\{\mathbf{h}_i\right\}_{i=1}^{N_p}$. Meanwhile, we represent the whole phrase using the last hidden vector, denoted as $\mathbf{h}^e_i$.
Second, we input the initial word representations and the hidden vectors to the self-attention modules $\operatorname{F}_{\rm attn}^{\rm app}$, $\operatorname{F}_{\rm attn}^{\rm pos}$ and $\operatorname{F}_{\rm attn}^{\rm spo}$ respectively, and get corresponding outputs.
Computation in self-attention modules is as follows:
\begin{equation}
	\resizebox{0.95\hsize}{!}{$
	\operatorname{F}_{\rm attn}^{M}\left(\left\{\mathbf{w}_i\right\}_{i=1}^{N_p},\left\{\mathbf{h}_i\right\}_{i=1}^{N_p}\right)=\sum_{i = 1}^{N_p} \frac{\exp\left(\mathbf{W}_M^T\mathbf{h}_i\right)}{\sum_{i = 1}^{N_p} \exp\left(\mathbf{W}_M^T\mathbf{h}_i\right) }\mathbf{w}_i$}
\end{equation}
where $\mathbf{W}_M^T$ is the trainable parameters of the module $\operatorname{F}_{\rm attn}^{M}$, and $M \in \left\{\rm app,pos,spo\right\}$.
For each entity phrase, we obtain an appearance feature $\mathbf{e}_{i}$ from $\operatorname{F}_{\rm attn}^{\rm app}$ and a location feature $\mathbf{l}_{i}^e$ from $\operatorname{F}_{\rm attn}^{\rm pos}$. Its linguistic feature from the LSTM is $\mathbf{h}^e_i$.
For each linguistic relation $r_{ij} \in \mathcal{R}$, its feature is denoted as $\mathbf{r}^e_{ij}$, which is achieved from $\operatorname{F}_{\rm attn}^{\rm spo}$ and encodes the words from the corresponding SPO phrase.


\subsection{\textbf{Details for Propagation}}\label{sec:sim}

We use some functions to build the correspondences between multi-modal entities and relations in Sec.~\ref{sec:program}. Their detailed implementations and explanation are shown as follows.
\paragraph{Similarity computation.} 
The similarity functions $\operatorname{F_{sim}}$ for entities and relations are implemented differently ($\operatorname{F_{sim}^{ent}}$ for entities and $\operatorname{F_{sim}^{rel}}$ for relations, $\operatorname{F_{sim}} \in \{\operatorname{F_{sim}^{ent}},\operatorname{F_{sim}^{rel}}\}$).
Inspired by SGRAF \cite{diao2021similarity}, we define similarity function $\operatorname{F_{sim}^{ent}}$ (in Eq.~\ref{eq:ent-sim}) of two vectors $\mathbf{x}$ and $\mathbf{y}$ as follows:
\begin{equation}
	\operatorname{F_{trf}}(\mathbf{x})=\mathrm{L}2 \mathrm{Norm} \left(\mathrm{MLP}\left(\mathbf{x}\right)\right)
\end{equation}
\begin{equation}
	\resizebox{0.95\hsize}{!}{$
\operatorname{F_{sim}^{ent}}(\mathbf{x}, \mathbf{y})=
\tanh\left(\mathbf{W}_{\rm eval}^{T}\sigma
\left(\frac{\mathbf{W}_{\rm sim}^{T}|\operatorname{F_{trf}}(\mathbf{x})-\operatorname{F_{trf}}(\mathbf{y})|^{2}}
{\left\|\mathbf{W}_{\rm sim}^{T}|\operatorname{F_{trf}}(\mathbf{x})-\operatorname{F_{trf}}(\mathbf{y})|^{2}\right\|_{2}}\right)\right), $}
\end{equation}
where $\operatorname{F_{trf}}$ represents the feature transformation, $\mathrm{L}2 \mathrm{Norm}$ is the L2 normalization, and all the MLPs are multi-layer perceptrons with ReLU activation. 
$\sigma$ represents ReLU function, $\mathbf{W}_{\rm eval}$ projects the vector similarity to a scalar value, and $\tanh$ is for normalizing the output scalar to the range of $\left[-1,1\right]$. The similarities $b_{ij}^{\rm app}$ and $b_{ij}^{\rm pos}$ for the appearence and spatial location space respectively are obtained from $\operatorname{F_{sim}^{ent}}$. The vector similarity has been shown to be more expressive than cosine similarity function.

The similarity function $\operatorname{F_{sim}^{rel}}$ (in Eq.~\ref{eq:fuseprod}) focusing on relations is implemented with the conventional cosine similarity in order to reduce the computational complexity:
\begin{equation}
\begin{aligned}
	[\mathbf{A}_{ij}]_{kl} &=\sigma\left(\left\langle\operatorname{F_{trf}}\left(\mathbf{r}^e_{ij}\right), \operatorname{F_{trf}}(\mathbf{r}^o_{kl})\right\rangle\right).
\end{aligned}
\end{equation}

\paragraph{Normalization.}
Another function for normalization $\operatorname{F_{norm}}$ divides all the values with the maximum absolute value if the maximum absolute value is larger than 1. If the final range is $[-1, 1]$ (in local correspondences), it will then linearly maps the range to $[0,1]$.
\paragraph{Independent parameters.}
Bottom-up and top-down propagations have different $\beta^{\rm rel}$ with independent parameters $ \mathbf{W_{\rm rel}}$ in Eq.~\ref{eq:gateprod}. The reasoning also learns different parameters $\mathbf{W_{\rm rel}}$ for grounding and matching, but shares the same local correspondences, i.e. the same parameters in $\operatorname{F_{sim}}$. Empirical results show this necessity because they focus on different aspects of the reasoning results. To be specific, the matching task emphasizes the global alignment representation, but the grounding task focuses more on the local alignment variance. 




\paragraph{Belief selection in the propagation for matching.}
Compared to the grounding, the matching task typically requires reasoning on the global representations of correspondences, rather than local variances on the belief. We apply a pruning strategy that only chooses top $K$ visual objects with the highest similarity scores to be the assignment space for each linguistic entity, getting $\mathbf{b}_{i} \in \mathbb{R}^K$ for all $i \in \left[1,N_e\right]$ in the matching propagation. Additionally, the belief vectors are sorted for computing the confidence score in Sec.~\ref{sec:program}. 
The pruning selects the most representative similarities in each correspondence feature, and sorting may make the confidence computing more sensitive to the sharpness of the belief vector. The relation correspondences are also pruned according to the node pruning results.

\section{Box Regression} \label{sec:regression}
We adopt detected object proposals for the generalized grounding, which could be a performance bottleneck due to their inaccurate localization. Consequently, we append an additional regression head to RCRN, UNITER and FGVE in order to refine the proposal locations.

We fuse visual and language features and apply an MLP to compute offsets for refining the proposal coordinates:
\begin{equation}
    \mathbf{\delta}= \operatorname{MLP_{\rm regress}} \left(\left[\mathbf{f}_{i}^O,\mathbf{f}_{i}^{\rm pos},\mathbf{f}_{\rm global}^L\right]\right)\\
\end{equation}
where $\mathbf{f}_{i}^O $, $\mathbf{f}_{i}^{\rm pos} $ and $\mathbf{f}_{\rm global}^L$ represents the predicted visual region representation, location feature of the predicted region and the global representation of language expression respectively. $\delta \in \mathbb{R}^4$ corresponds to the offsets for refining the region proposal.

For instance, in RCRN, $\mathbf{f}_{i}^O$ is the feature $\mathbf{o}_{i}$ of the predicted box, $\mathbf{f}_{i}^{\rm pos} $ is the 5-dimensional location feature $\mathbf{l}_{i}^O$ for the predicted region, and $\mathbf{f}_{\rm global}^L$ is the the mean of all SPO representations. 

During training, we use a smoothed $L_1$ regression loss to penalize the difference between $\delta$ and the ground-truth offsets $\delta^{\ast}$. 
\begin{equation}
    \mathcal{L }_{\rm reg}=\operatorname{smooth}_{L_1}\left(\delta^{\ast},\delta\right)
\end{equation}
where $\delta^{\ast}$ is the difference between the coordinates of a ground-truth bounding box and that of a predicted box.

\section{Dataset construction} \label{sec:data}
\subsection{Data Generation}
We construct our dataset {GITM-MR} by a generation program to create mismatch expressions from the corresponding original expressions by partial replacement. First, we identify all the relation phrases in the expressions of the Ref-Reasoning dataset by using an off-the-shelf parser \cite{schuster2015generating}. 
After that, we manually select a subset of 27 commonly-occurred relations, and assign some relations acceptable to similar contexts but with different semantics, for each relation in the subset, as their replacement candidates. For example, we assign ``carry" as an candidate for ``wear", and ``to the left of" as an candidate for ``to the right of". Utilizing human annotations as such significantly reduces the linguistic bias and false negative cases on the generated expressions. Finally, we replace one relation in each expression by a candidate to construct a mismatch expression. The program keeps the relation phrase set and the replaced relation for each expression as the labels for MRR evaluation.

The following aspects are considered additionally to control the overall quality of the generated mismatched expressions.

\paragraph{Mismatched relation diversity.}
In the real-world scenario, the mismatched expressions should be various as the matched expressions, rather than showing certain patterns which are easy to identify. To keep the diversity, we assign multiple substitutions for each relation. The full replacement candidate list is shown in Tab.~\ref{tb:rpl} for the reference. We randomly select the substitution when replacing a relation.

\begin{table}[t!]
	\centering
	\vspace{0.5em}
	\renewcommand\arraystretch{1.0}
	\resizebox{0.45\textwidth}{!}{
			
			\begin{tabular}{c|c}
				\toprule[1pt]
				\textbf{Relation} & \textbf{Substitutions} \\
				\hline
				to the left of & to the right of, in front of, behind \\
				\hline
				to the right of & to the left of, in front of, behind \\
				\hline
				wearing & holding, carrying, looking at \\
				\hline
				in front of & behind, to the right of, to the left of \\
				\hline
				behind & in front of, to the right of, to the left of \\
				\hline
				holding & looking at, behind \\
				\hline
				on top of & on the side of, near, next to \\
				\hline
				above & near, next to, behind, in front of \\
				\hline
				below & near, next to, in front of, behind \\
				\hline
				sitting on & near, next to, behind, in front of, to the left of, to the right of \\
				\hline
				next to & in front of, behind \\
				\hline
				carrying & looking at, behind \\
				\hline
				inside & near, next to, behind, in front of, to the left of, to the right of \\
				\hline
				under & above, on top of, next to, in front of, behind \\
				\hline
				standing on & walking on \\
				\hline
				walking on & standing on \\
				\hline
				eating & holding, looking at, behind \\
				\hline
				standing in & walking in \\
				\hline
				playing with & standing by, behind \\
				\hline
				walking in & standing in \\
				\hline
				riding & behind, in front of, to the left of, to the right of \\
				\hline
				riding on & behind, in front of, to the left of, to the right of \\
				\hline
				playing & standing by, behind \\
				\hline
				walking down & standing on \\
				\hline
				throwing & holding, behind \\
				\hline
				standing behind & standing in front of \\
				\hline
				standing in front of & standing behind \\
				\bottomrule[1pt]      
			\end{tabular}
		}
	\caption{\small{
			The full substitution candidate list of the selected relations.}}
	\vspace{0.5em}

\label{tb:rpl}
\end{table}

\paragraph{Linguistic bias.}
We reject the substitutions that violate the following rules to further control the linguistic bias:
\begin{enumerate}
	\item The generated mismatched expression should have close perplexity in the pre-trained language model BERT \cite{kenton2019bert} with the original matched expression.
	\item After the substitution, the relation phrase occurrence distribution in matched and mismatched expressions should be close.
\end{enumerate}

\begin{figure}[htbp]
	\centering
	\includegraphics[width=0.45\textwidth]{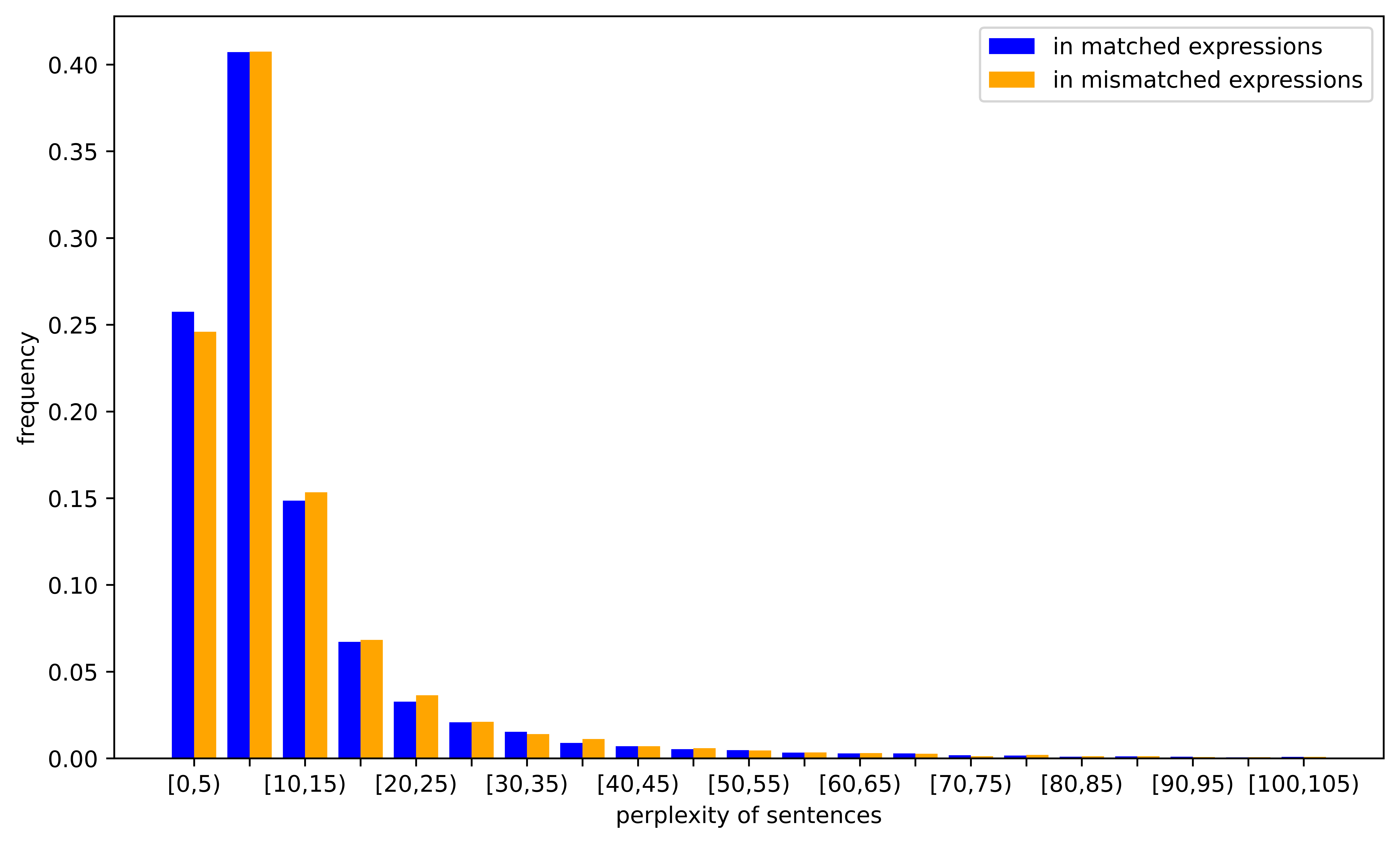}
	\caption{Sentence perplexity distributions on the validation set. Only the perplexities ranging from 0 to 105 are shown for the ease of reading.}
	\label{fig:ppl}
\end{figure}

\begin{figure}[htbp]
	\centering
	\includegraphics[width=0.45\textwidth]{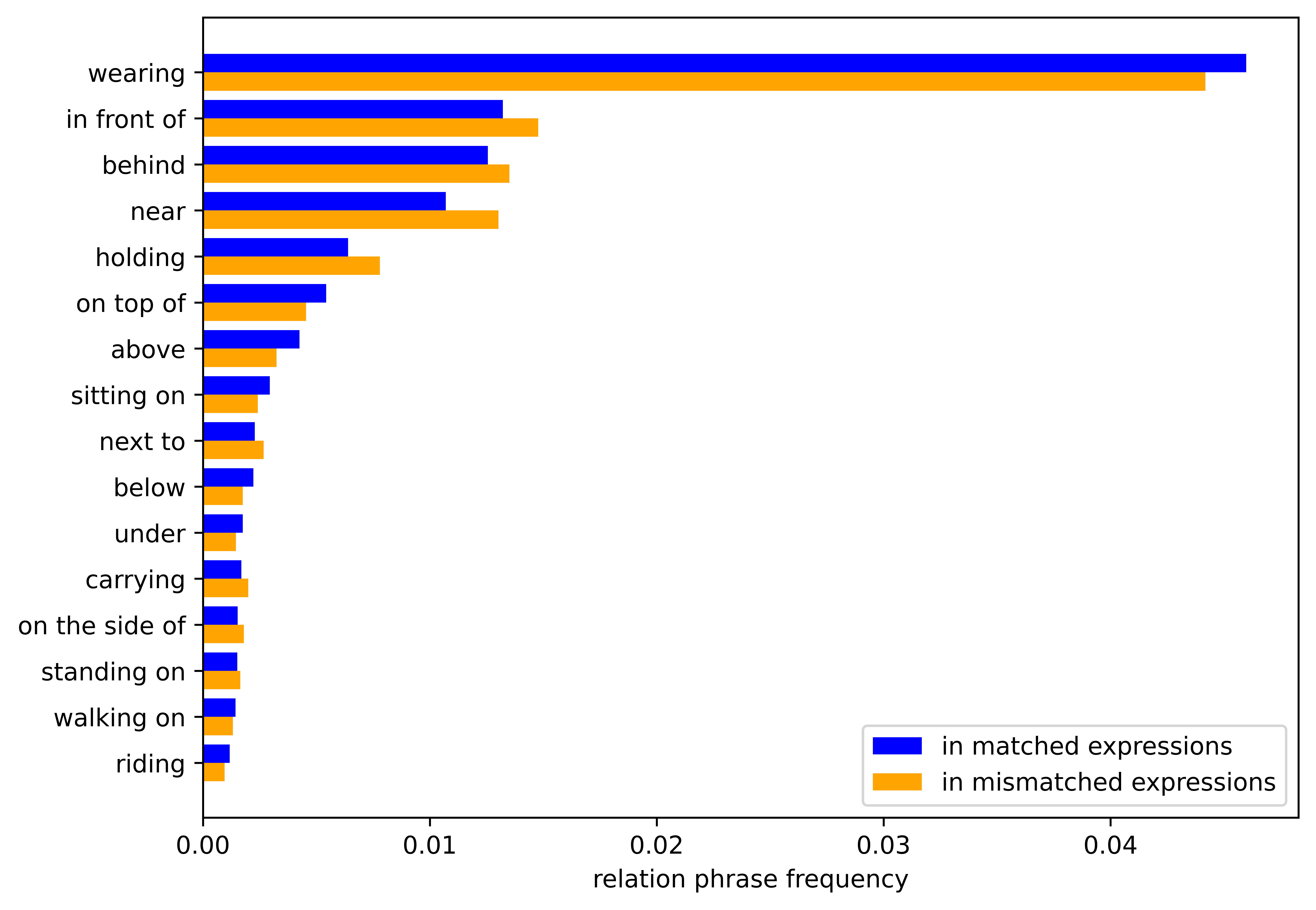}
	\caption{Relation frequencies on the validation set. Only the body part of the relations are shown for the ease of reading.}
	\label{fig:rel_f}
\end{figure}

We count the distribution of sentence perplexity in the range of 0 to 105 in Fig.~\ref{fig:ppl}, and the frequencies from the main body of the relation phrases belonging to the substitution list in Fig.~\ref{fig:rel_f}. The statistics of matched and mismatched expressions in the validation set are shown by the bars with different colors. The results demonstrate the high similarity between matched and mismatched expressions.

Furthermore, we generate several candidate datasets, and then train language-only transformers to do binary classification on each dataset. The lower accuracy on the task implies the lower linguistic bias on the dataset. The candidate dataset with the lowest classification accuracy is chosen as the final GITM-MR dataset. 

\paragraph{False mismatch.}
There still exist image-text pairs where a relational phrase in an expression is replaced, but the entire expression still perfectly describes an object in the image. We call these false mismatched cases. To avoid those unwanted cases from hurting the quality of testing, we hired annotators from Stardust\footnote{\href{https://stardust.ai/}{https://stardust.ai/}} to label the mismatch cases in the test set, indicating whether the constructed mismatched sentence description in each case can match an object in the image. In the end, we only keep the cases that the annotators identified as mismatch in the test set. The final test set contains only the reserved mismatch cases and their corresponding matched cases. We provide more details of the manual curation in Sec.~\ref{sec:manual}

\subsection{More Details on Manual Data Curation}\label{sec:manual}
\begin{figure*}[htbp]
	\centering
	\includegraphics[width=0.8\textwidth]{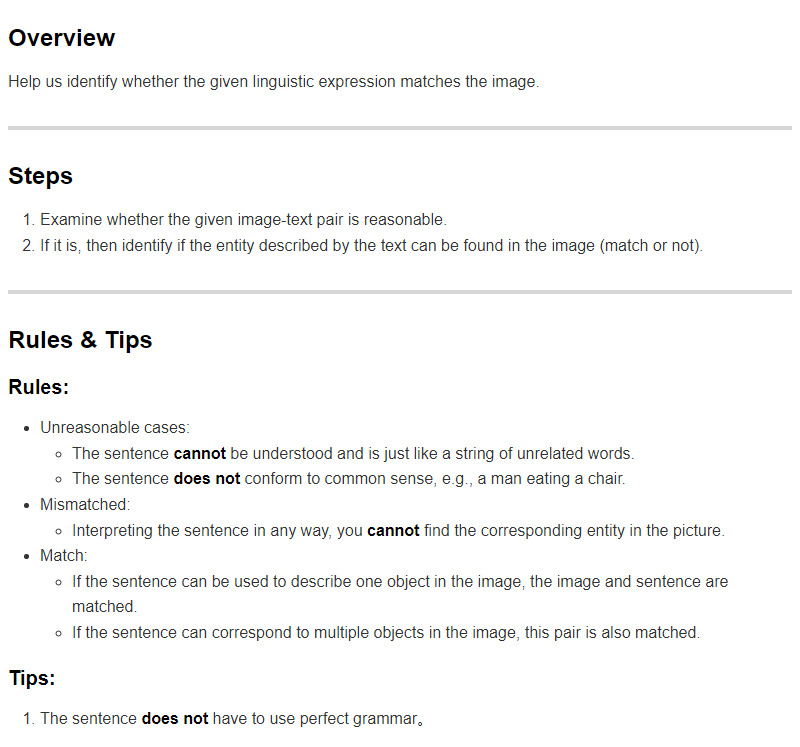}
	\caption{A screen-shot of instructions provided to the annotators to filter the false mismatched cases.}
	\label{fig:instru}
\end{figure*}
\begin{figure*}[htbp]
	\centering
	\includegraphics[width=0.9\textwidth]{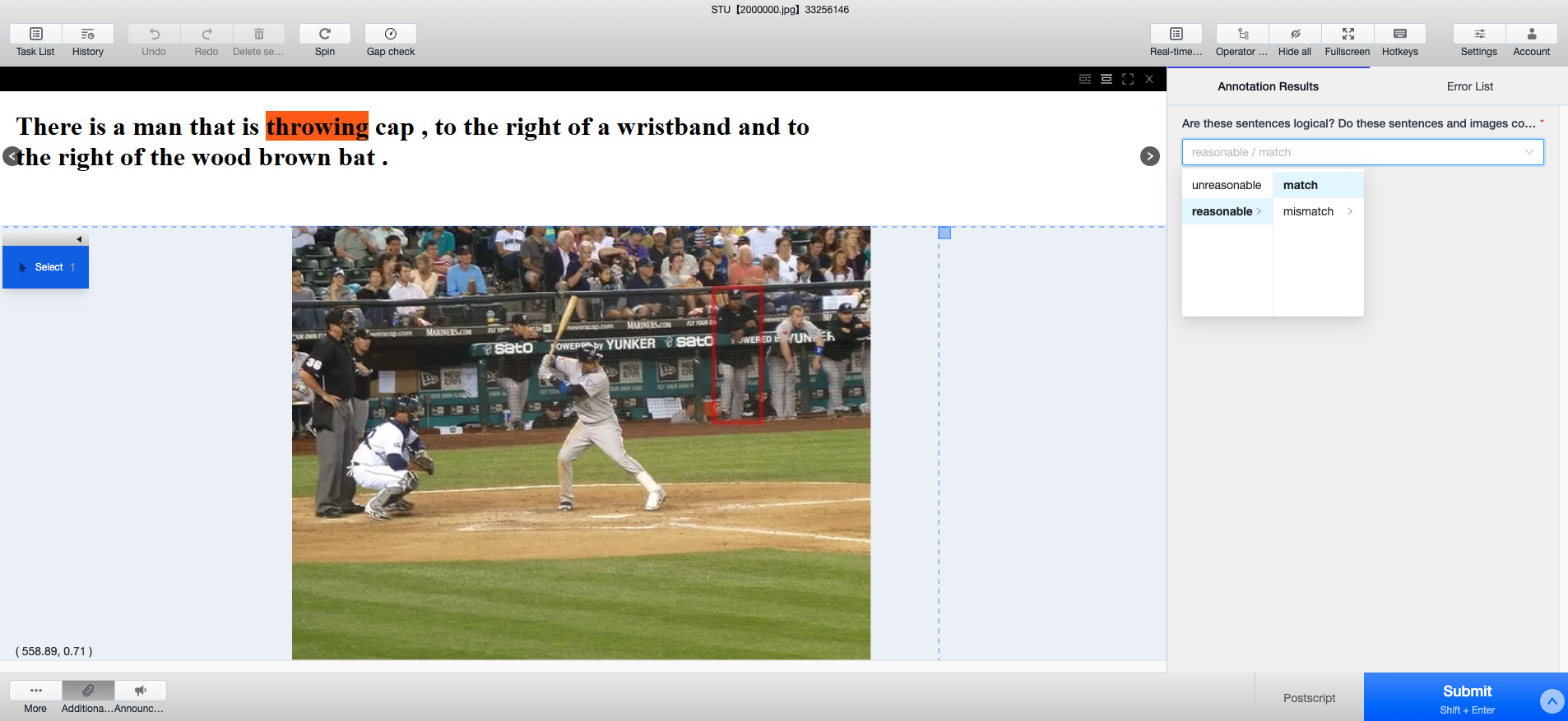}
	\caption{A screen-shot of user interface for the manual data curation.}
	\label{fig:user}
\end{figure*}
\paragraph{User interface.}
We provide instructions for labeling given to workers in Fig.~\ref{fig:instru}. 
The user interface of data curation on the StarDust platform is shown in Fig.~\ref{fig:user}. 
The langauge expression is shown in the top part and the image is shown in the left. In the right part, the tester is asked with two question.
In the first step, the worker needs to examine whether the sentence is reasonable. If the answer at first step is positive, then the worker should identify if the sentence describe an object in the image at the second step.

Since the dataset involves relatively complex image scenes and sentence structures, we annotate one box in each image to help workers make a decision at the second step. The annotated box is the ground truth box in the original grounding dataset, i.e. Ref-Reasoning~\cite{yang2020graph}. Because the sentence in the mismatched case is obtained by replacing one relation in the sentence from the original dataset, it is possible that the sentence constructed with substitution still points to the object corresponding to the original sentence. We also color the replacement in the sentence to help workers focus more on the important parts of sentences. To avoid the workers from biased labeling, we didn't tell them the construction of mismatched cases.

\begin{figure*} [ht]
	\centering
	
	\subfigure[]{
		\includegraphics[width=5.5cm]{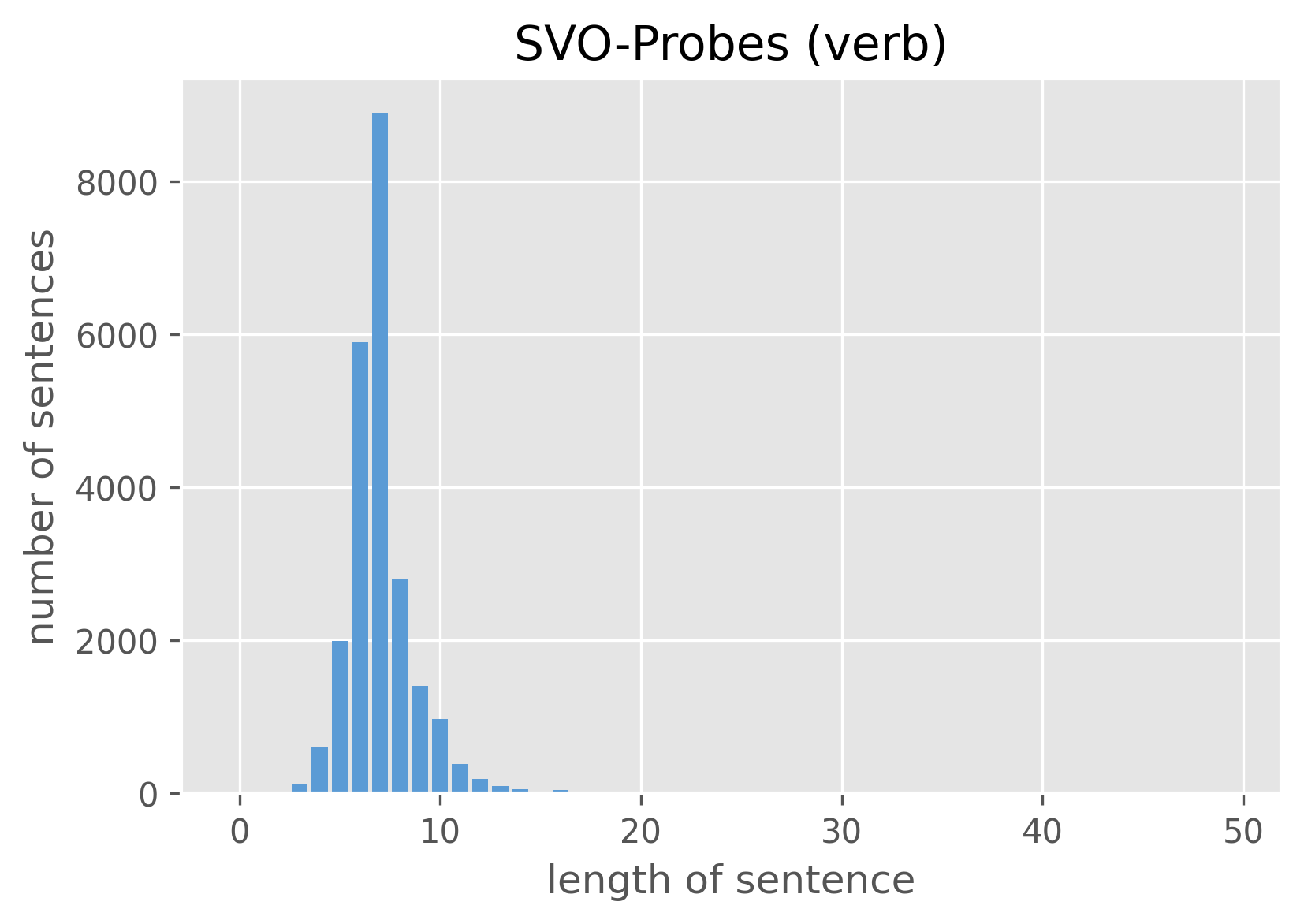}
	}
	\subfigure[]{
		\includegraphics[width=5.5cm]{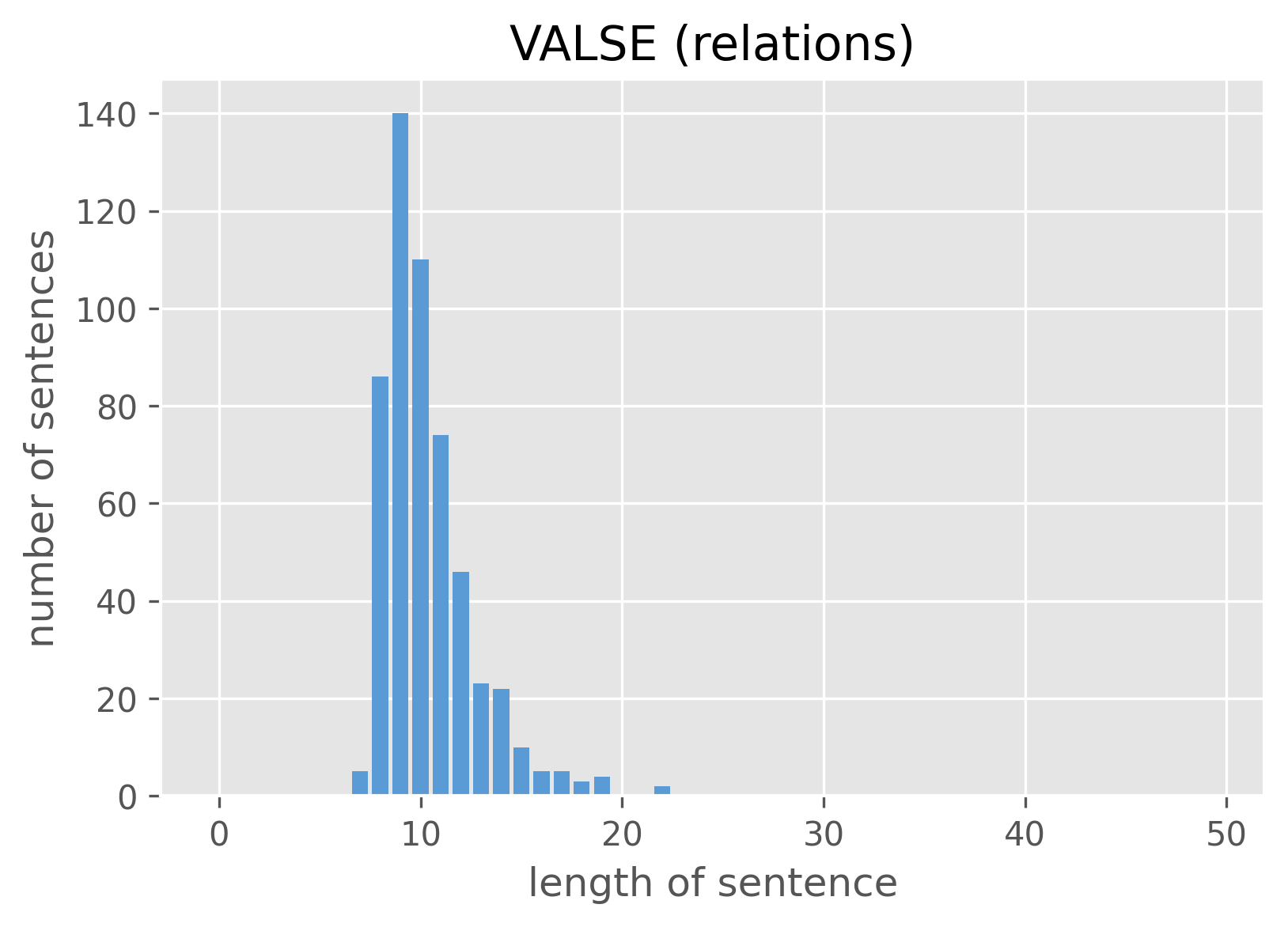}
	}
	\quad
	\subfigure[]{
	\includegraphics[width=5.5cm]{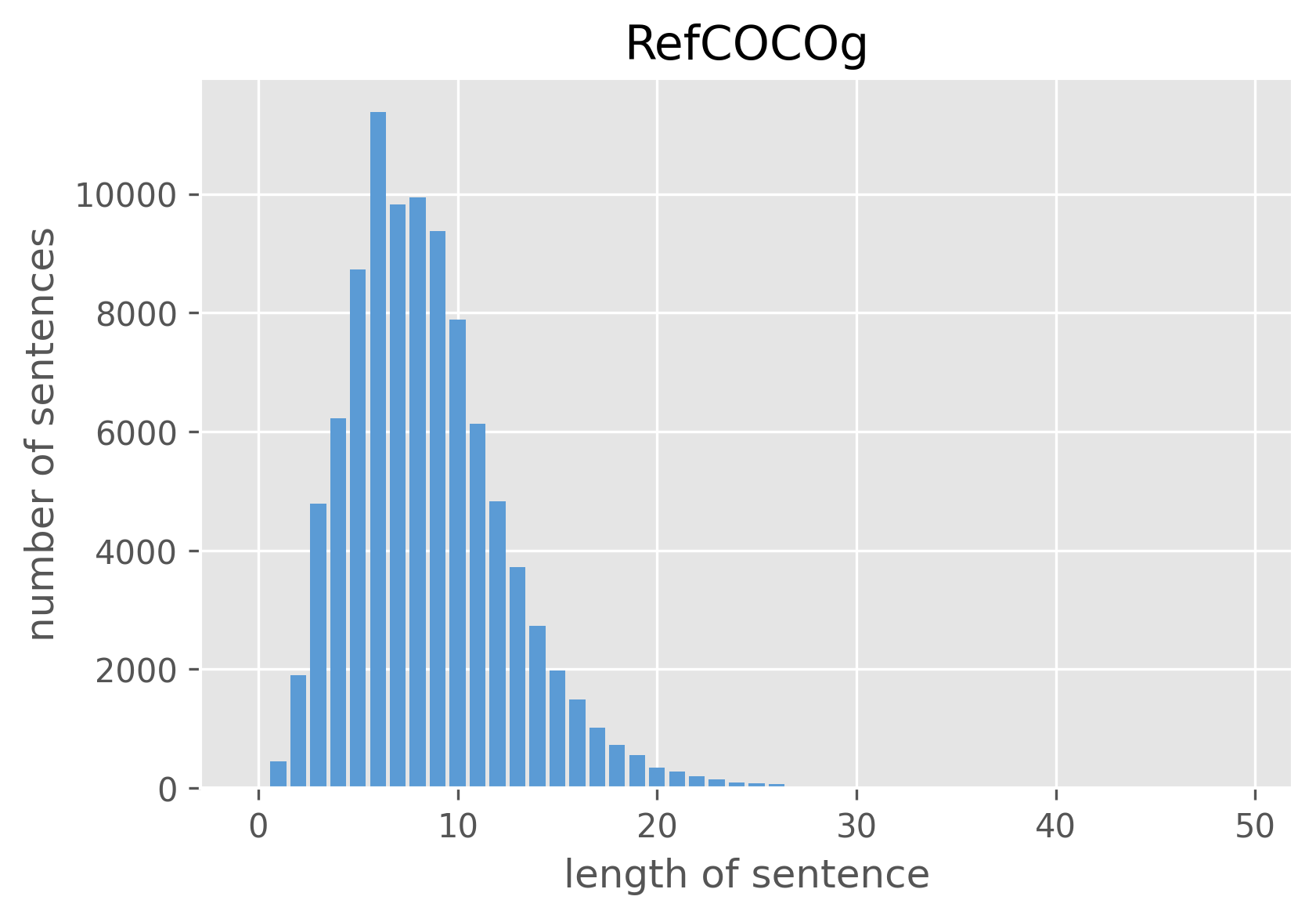}
	}
	\subfigure[]{
	\includegraphics[width=5.5cm]{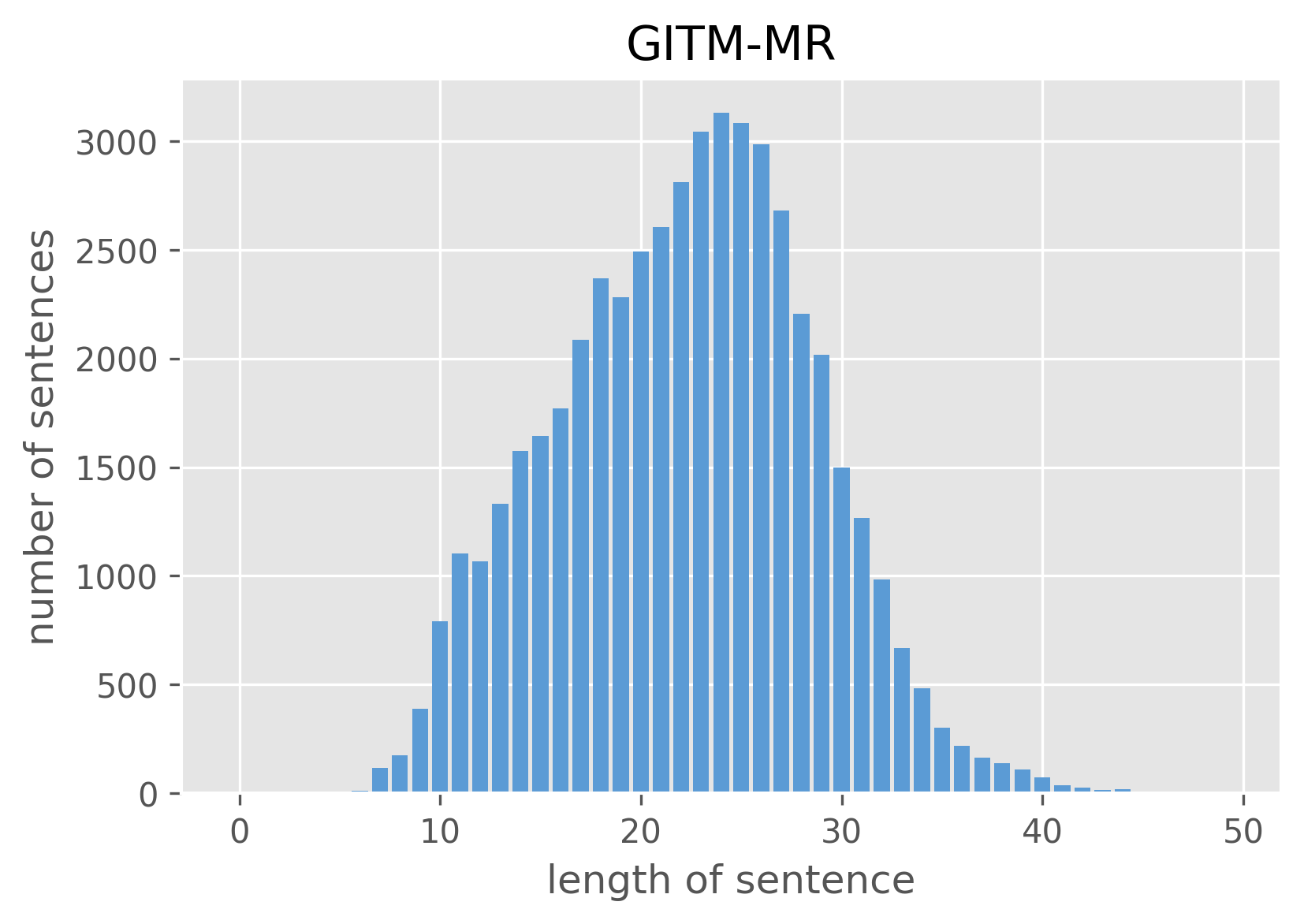}
	}
	\caption{The distribution of sentence length in different datasets. Overall, our dataset has longer sentences than other datasets. 
	In terms of details, the length of most sentences in other datasets is within 16, which corresponds to the standard for defining our OOD setting in Sec.~\ref{sec:app-data}.}
	\label{fig:data_dist} 
\end{figure*}

\paragraph{Job setting.}
Each StarDust annotator maintains a job approval rate based on their performance on previous jobs. We invite only experienced annotators whose job approval rate is equal to or greater than 98\%.
Also, we hire three independent annotators for each job and aggregate their annotations for final decision.

\paragraph{Quality control.}
In the labeling period, Stardust submitted the data labeled everyday to their platform. We checked the data after their each submission. For the manual labels that we disagreed with, we gave the reason for judgment and rejected the samples for workers to re-label. Each day, annotators also maintained a document that stores data that they considered ambiguous, and asked questions about the labeling process. We answered their questions every day in the document, and unified the labeling standards for some details.

\paragraph{Result.} 
Finally, we only keep the image-text pairs where at least two testers can't find a relative object in the image according to the text. The reserved mismatch subset contains 2046 images and 5616 expressions.

\subsection{Comparison with Other Datasets}\label{sec:compare data}

\begin{table}[t]
	\centering
	\renewcommand\arraystretch{1.3}
	\resizebox{0.4\textwidth}{!}{\begin{tabular}{c|ccc}
	\toprule[1pt]
	\textbf{Dataset} & \textbf{\makecell{Sentence Length\\(mean/median)}} & \textbf{\makecell{Average Number\\ of Entities}} & \textbf{\makecell{Perplexity\\(pos/neg)}} \\ 
	\hline\hline
	RefCOCO          & 3.50/3                               & 1.21                             & -                   \\
	RefCOCO+         & 3.53/3                               & 1.10                             & -                   \\
	RefCOCOg         & 8.46/8                               & 2.29                             & -                   \\
	SVO-Probes (verb)         & 6.21/6                               & 2.01                             & -                   \\
	VALSE (relations)        & 10.36/10                             & 2.73                             & 25.03/34.45         \\
	GITM-MR          & 22.22/23                             & 3.55                             & 12.85/13.20         \\ 
	\bottomrule[1pt]
	\end{tabular}
	}
	\caption{Comparsion with other datasets from the aspects of sentence length, number of entities in each sentence and perplexity gap between sentences from positive and negative cases.}
	\label{tab:compare dataset}
\end{table}

In this section, we compare our proposed benchmark GITM-MR with existing datasets of grounding or matching task, showing the advantage of our GITM-MR on sentence complexity and sentence rationality. 
In Tab.~\ref{tab:compare dataset}, RefCOCO~\cite{nagaraja2016modeling}, RefCOCO+~\cite{nagaraja2016modeling} and RefCOCOg~\cite{mao2016generation} are three datasets for the visual grounding task. 
SVO-Probe~\cite{hendricks2021probing} is a dataset designed to test pre-trained VL models' understanding of verbs, subjects and objects. We make statistics on its subset involving negative verbs.
VALSE~\cite{parcalabescu2022valse} is a benchmark aimed at gauging the sensitivity of pre-trained VL models to \textit{foiled} instances, the statistics in this table refers to one VALSE subset involving spatial relations. 
\paragraph{Sentence complexity.} 
We compare the sentence complexity from two aspects: sentence length and number of entities in each sentence. Firstly, Tab.~\ref{tab:compare dataset} and Fig.~\ref{fig:data_dist} shows that the average sentence length of GITM-MR is much longer than other datasets. 
Secondly, in Tab.~\ref{tab:compare dataset}, we compared the number of parsed entities in each dataset, and GITM has the highest average number of entities in each sentence. More entities in one sentence means more relations exist, which makes the reasoning process involved in grounding and matching tasks more complicated or leads to longer reasoning paths.

\paragraph{Sentence rationality.} 
In Tab.~\ref{tab:compare dataset}, we measure the rationality of the constructed negative sentences by comparing the perplexity difference between the positive and negative cases in each dataset, following VALSE. Compared with VALSE (split on relation substitution), the perplexity gap between the positive and negative cases in our GITM-MR is much smaller, which means the constructed negative examples have a smaller plausibility bias, as described in \cite{parcalabescu2022valse}.


\section{Experiments} \label{sec:exp}
\subsection{Setup}
\subsubsection{Training Details on RCRN}
To prevent numerical instability, we apply several techniques when implementing and training RCRN. 
(1) We implement the multiplication in Eq.~\ref{eq:fuseprod} and 6 using log space transformation. We firstly substitute production with summation in log space, and then transform the value back by the exponential function. 
(2) We initialize the learning by training only on grounding task from scratch to learn transformation layers in similarity functions first. Empirical results show that these learned layers are helpful for later multi-task joint learning.
(3) We cut off the gradients on all the input of $\beta_i$'s for stable training. 

\subsubsection{VL Pre-trained Models for GITM-MR} \label{sec:UNITER}
As referred in Sec.~\ref{sec:exp setup}, we modify UNITER, TCL, FIBER and FGVE to complete the three substasks of GITM-MR. In terms of the modification for them, we use the pretrained components from original models, and add two subtask heads for grounding and matching. 

\paragraph{UNITER}
For the matching task, we extract the representation of [CLS] token as the joint representation of the input image-text pair, and then feed it into an FC layer and a sigmoid function to predict a score between 0 and 1. We apply the binary cross-entropy loss on the matching scores for optimization. 
In order to complete the grounding task, we add a MLP layer on top of the region token outputs from transformer layers to compute the alignment score between the expression and each visual region. Then we apply cross-entropy loss on the normalized alignment scores.

\paragraph{UNITER+MIL}
To allow the pre-trained models to simultaneously complete the MRR task, we conduct a multi-instance learning strategy following \cite{wang2019comparison}, creating the baseline UNITER+MIL. Concretely, for the matching task, we firstly take the textual token outputs from Transformer, and feed it into the FC layer (matching head). Thus we get a local matching score for each textual token. Then we get minimum local matching score instead of the score for [CLS] token as the global prediction to compute the matching loss. 
For the MRR task, we compute a matching score for each relation in the language expression assisted by the language parser. The score for a relation is the average of local matching scores for words in this relation. Finally, the relation earning the lowest score is the predicted mismatched relation. 

\paragraph{TCL}
TCL is modified for the GITM-MR task similarly. Different from UNITER which using image token representations for grounding, TCL is a one-stage pretrained model without output representation for each proposal in the image. So TCL can only complete the matching task in GITM-MR.

\begin{table*}[ht]
	\centering
	\renewcommand\arraystretch{1.3}
	\resizebox{0.8\textwidth}{!}{
		\begin{tabular}{cc|cc|cc|cc}
			\toprule[1pt]
			&  & \multicolumn{2}{|c|}{\textbf{Full Test}}& \multicolumn{2}{|c|}{\textbf{In-Distribution}}& \multicolumn{2}{|c}{\textbf{Out-of-Distribution}} \\
			\textbf{Training Set} &\textbf{Method}    &\textbf{Grounding\%}  &\textbf{MRR\%}   &\textbf{Grounding\%}  &\textbf{MRR\%}  &\textbf{Grounding\%}  &\textbf{MRR\%}\\ 
			
			\hline\hline
			Train-Len16           
			& UNITER       	&40.21	&-		&52.97	&-		&37.50	&-\\
			& UNITER+MIL   	&41.62	&60.77		&54.55	&89.96		&38.88	&54.53\\
			& FGVE+MAX	   		&-	    &49.57		&-	&79.24  &-	&43.22\\
			& FIBER	        &28.51 &- &\textbf{57.24} &- & 22.41 &-\\
			& RCRN(Ours)   	&\textbf{42.06} &\textbf{72.08}		&54.43	&\textbf{90.74}		&\textbf{39.43}	&\textbf{68.09}	\\
			\hline
			Train-Len11           
			& UNITER       	&35.04	&-		&45.81	&-		&34.59	&-\\
			& UNITER+MIL   	&32.27 	&52.67		&41.87	&96.60		&31.87	&50.82 \\
			& FGVE+MAX		&- 	&49.71		&-	&97.57		&-	&47.68 \\
			& FIBER	        &25.72 & - & \textbf{51.23} & - & 24.66 & - \\
			& RCRN(Ours)    &\textbf{38.16}&\textbf{65.55}  &45.81	&\textbf{98.54}		&\textbf{37.85}	   	&\textbf{64.16}\\
			\bottomrule[1pt]
		\end{tabular}
	}
	\caption{\small{Oracle experiment results on the GITM-MR dataset. In this experiment, the computation of grounding and MRR accuracies are independent from the matching results.
	}}
	\label{tb:generalize oracle}
\end{table*}

\paragraph{FIBER}
We use the fine-grained pre-trained FIBER to complete GITM-MR. For the grounding task, the OD-head from FIBER is preserved. For the matching task, following the coarse-grained pre-trained FIBER, we concatenate the two global representations from both the visual and language backbone, and feed them into a MLP layer to get the matching score. We also tried to use the coarse-grained pre-trained FIBER to complete the matching task, but it struggles to get an accuracy of just over 50\%. 

\paragraph{FGVE}
To modify FGVE for the MRR task, we use Fast Align Algorithm Aligner\footnote{\href{https://amrlib.readthedocs.io/en/latest/faa\_aligner/}{https://amrlib.readthedocs.io/en/latest/faa\_aligner/}} tool to align the knowledge elements (KE) with tokens in the sentence, and then maximally pool the mismatch score predicted on each KE to aligned tokens. Once the score on each token is obtained, the mismatch score on each relation phrases are computed as the maximum of their token scores, and the one with the highest mismatch score will be MRR prediction. The multi-instance learning losses are preserved and modified for our binary classification matching task.

\subsubsection{Data Partition Criterion for OOD setting}\label{sec:app-data}


As mentioned in Sec.~\ref{sec:setup}, we construct training sets including only simple sentences. The models are trained on the simple training set, and evaluated on test set containing both simple and complex sentences. Next, we describe the criterion of selecting simple image-sentence pairs for training from two aspects: number of entities and sentence length. 

\paragraph{Number of entities.}
The complexity of a sentence is proportional to the number of entities contained in it. Smaller number of entities represents lower sentence complexity. However, the number of entities obtained from the off-the-shelf parser is not accurate. We only get a part of ground-truth numbers of entities in val set. Due to the fact that the length of a sentence is also positively correlated with the number of entities, we use the ground-truth numbers of entities in val set to infer appropriate sentence lengths as dividing lines between the simple and the complex.

Following this idea, we first draw a histogram (Fig.~\ref{fig:node and length} in the main paper) to observe the distribution of sentence length and number of entities in the validation set. Second, we choose appropriate sentence lengths for division according to the histogram. Finally, the selected sentence lengths are applied to select training set containing simple sentences.

The reason for choosing sentence lengths 11 and 16 as dividing lines is shown as follows. In Fig.~\ref{fig:node and length} in the main paper, when sentence length is shorter than 11, most sentences contain only two entities, and a few sentences contain three entities. In sentences which have length less than 16, two or three entities are mainly included, and there still exists small number of sentences with four entities.

\paragraph{Sentence length.}
As shown in Fig.~\ref{fig:data_dist}, in the common datasets of matching and grounding tasks, most sentences are within 16 in length. This shows that sentences shorter than 16 are easy to collect for training, but longer sentences are more difficult to obtain. So it is reasonable to choose Train-Len16 as an OOD scenario.
Moreover, Train-Len11 is a more challenging evaluation setting, because the simpler (shorter) the sentences in the training set, the higher the generalization ability of the model is required.

\subsubsection{Discussion on Length Generalization}
The length generalization of models serves as a crucial test of their ability to understand relations. In the GITM-MR task, length generalization can be viewed as a proxy for relation number generalization. As longer sentences typically contain more relations, they require more complex reasoning paths to identify the referents, much like the concept of reasoning hops in VQA tasks, where the number of reasoning steps required to answer a question reflects the level of reasoning needed. Thus, the ability of models to generalize to longer sentences with more relations demonstrates their capability to handle complex relation understanding tasks and to perform reasoning across multiple entities and relations.

\subsection{Results}
\subsubsection{Oracle Results}
As mentioned in the dicussion in Sec.~\ref{sec:result}, we investigate models' performance on grounding and mismatch relation reasoning by evaluting for results of these two tasks independently, without considering the matching prediction. The results in Tab.~\ref{tb:generalize oracle} shows that our \M~ achieves high accuracies on both subtasks in the different training setups. Particularly, \M~ outperforms other models on the MRR task, and the gap is especially notable in the OOD test. This demonstrates our models' strong ability to complete the grounding and MRR task, which requires models to learn fine-level cross-modal alignments.

\subsubsection{Additional Results on Limited Training Data}
\begin{table}[tbp]
	\centering
	\vspace{0.5em}
	\renewcommand\arraystretch{1.3}
	\resizebox{0.45\textwidth}{!}{
		\begin{tabular}{c|ccc}
			\toprule[1pt]
			\multicolumn{1}{c|}{\textbf{Method}}                &\textbf{Matching\%} &\textbf{Grounding\%} &\textbf{MRR\%}  \\
			\hline\hline
			TCL                             & 52.14          & -              & -           \\
			UNITER                                 & 56.05    & 21.79                & -        \\
			UNITER+MIL      & 56.14                  & 20.37                 & \textbf{38.83}      \\
			FIBER & 54.01 & 23.95 & - \\
			\hline
			RCRN(ours)                                          & \textbf{60.97} & \textbf{29.96}   & 38.07\\
			\bottomrule[1pt]      	
		\end{tabular}
		
	}
	
	\caption{\small{
			Additional results trained on a subset of 5\% in-distribution training data. 
			The `-' denotes that the model is not applicable in corresponding subtask. 
			The results show that the proposed method outperform prior methods in limited data setting. }}
	\vspace{0.5em}
	
\label{tb:additional}
\end{table}
To further demonstrate the data efficiency of our \M~ in the limited data setting, we evaluate our result on an additional training setting. We uniformly sample 5\% training data from the original training set, and evaluate on the original test set. The result shown in Tab.~\ref{tb:additional} verifies the high performance of our method under limited training data setting again, and this superiority is independent of whether the maximum expression length is limited.

\subsubsection{Comparsion with Interpretable Models without VL Pre-training}

The proposed RCRN without the transformer layers (denoted as RCRN-T) can independently handle these three subtasks as a single unified model with interpretability.
In Tab.~\textcolor{red}{\ref{tb:lightweight}}, we compare our RCRN-T with several existing state-of-the-art interpretable models without vision-language pre-training for REG or image-text matching tasks on both GITM-MR and Ref-Reasoning~\cite{yang2020graph} dataset. 
The DGA and SGMN are grounding methods, and SGR is a ITM model. 
All these state-of-art models don't embrace appropriate design to accomplish the MRR task. 
For a fair comparison, all the models use the same visual object features and the same setting in word embedding.

\begin{table}[tbp]
	\centering
	\vspace{0.5em}
	\renewcommand\arraystretch{1.3}
	\resizebox{0.45\textwidth}{!}{
		\begin{tabular}{c|ccc|c|c}
			\toprule[1pt]
			\multicolumn{1}{c|}{} & \multicolumn{3}{|c|} {\textbf{GITM-MR}} & \multicolumn{1}{|c|}{\textbf{Ref-Reasoning}} & \\
			\multicolumn{1}{c|}{\textbf{Method}}                &\textbf{Mat} &\textbf{Grd} &\textbf{MRR}   & \textbf{Grd} &\textbf{\#Param}\\ 
			\hline\hline
			DGA\cite{yang2019dynamic}                             & -          & 36.49              & -           & 45.87 & 50.99M\\
			SGMN\cite{yang2020graph}                                 & -     & 42.79                & -           & 51.39 & 38.71M \\
			SGR\cite{diao2021similarity}      & 56.30                  & -                       & -           & - & 37.38M\\
			\hline
			Ours(RCRN-T)                                          & \textbf{67.12} & \textbf{55.64}   & \textbf{71.72}& \textbf{58.92} &\textbf{35.52M} \\
			\bottomrule[1pt]      	
		\end{tabular}
		
	}
	
	\caption{\small{
			Comparison with state-of-the-art interpretable models for REG and image-text matching tasks. 
			The `-' denotes that the model is not applicable in corresponding subtask. 
			Models are evaluated on ground-truth objects by following the setting in SGMN.}}
	\vspace{0.5em}
	
\label{tb:lightweight}
\end{table}

Tab.~\textcolor{red}{\ref{tb:lightweight}} shows that the proposed RCRN-T outperforms all existing state-of-the-art interpretable models under the same backbone setting and earns minimum number of parameters .
SGR achieves a relatively low accuracy 56.30\% on matching task, which shows that slight difference on the relations in sentences is hard to distinguish without special fine-level relation semantic learning design. DGA only learns a low-order language guided contextual representation for objects, and relatively fixed context modeling design limits SGMN's learning process. 
Compared with Ref-Reasoning, GITM-MR is more challenging on grounding because it has at least 2 entities in each expression, and it includes OOD scenarios in testing.

\subsubsection{Ablation on VLP Layers}
Tab.~\ref{tb:abl-vlp} shows the ablation study on the number of VLP layers used for generating the candidate representation. The results on the full test set show that our RCRN gets the best performance with 6 pre-trained VLP layers. 
Features from shallower layers may not have learned accurate enough cross-modal correspondence. Features from the next few layers may have overfitted to the finetuning tasks, and can't be refined through the propagation process easily.
\begin{table}[t]
	\centering
	   \renewcommand\arraystretch{1.3} 
	\resizebox{0.3\textwidth}{!}{
		\begin{tabular}{c|ccc}
			\toprule[1pt]
			VLP Layers & \textbf{Mat} & \textbf{Grd} & \textbf{MRR}\\ 
			\hline\hline
			 3 &62.28     & 26.88   & 52.97 \\ 
			 5 & 62.25	&28.05	&50.40 \\
			6 &\textbf{62.52}     & 26.71 &	\textbf{53.69} \\
			7 &62.18	&\textbf{29.08}	&47.83 \\
			9 & 62.07 &	26.43 &	52.99 \\
			12 & 62.12 & 27.13 & 50.71 \\
			\bottomrule[1pt]
		\end{tabular}
	}
	\caption{\small{Ablation study for VLP layers on GITM-MR validation set.}}
	\vspace{-5mm}
	\label{tb:abl-vlp}
	
\end{table}

\subsection{Visualization and Interpretability}

\subsubsection{Qualitative Results}

\begin{figure*}
	\centering
	\includegraphics[width=0.9\textwidth]{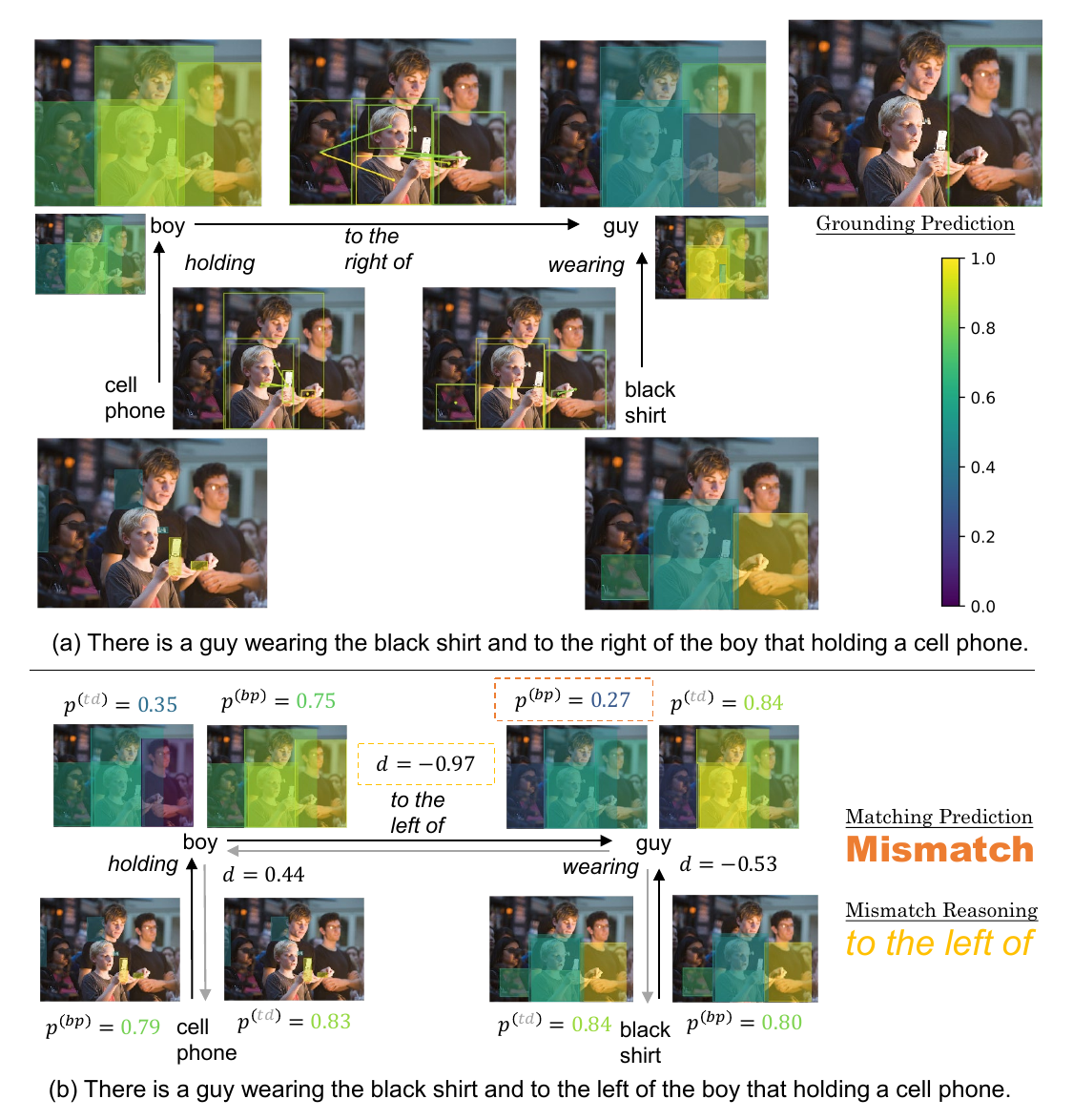}
	\vspace{1.0em}
	\caption{The visualization of the reasoning process of RCRN. The entity and relation correspondence maps are shown by color maps on boxes and box pairs respectively, where only those with top 5 correspondence scores for each phrase are included. Small color maps in (a) are the local correspondence maps and the large ones are the propagated contextualized maps. The boxes of parents in box pairs are dotted on their centers. Dashed boxes indicate the pooling results.}
	\label{fig:vis}
	\vspace{-1.5em}
\end{figure*}

As shown in Fig.~\ref{fig:vis}, we visualize a pair of matched and mismatched samples of our RCRN. Fig.~\ref{fig:vis}\textcolor{red}{(a)} shows the bottom-up propagation process followed by grounding prediction on a matched image-text pair, and Fig.~\ref{fig:vis}\textcolor{red}{(b)} presents the bi-directional propagation results with readout process for MRR on a mismatched pair. In Fig.~\ref{fig:vis}\textcolor{red}{(a)}, the propagation eliminates the ambiguities on entities under the guidance of explicit relation correspondences. Fig.~\ref{fig:vis}\textcolor{red}{(b)} illustrates that the intermediate reasons of the predictions can be traced. The sum of match confidence differences along the mismatched relation, denoted as $d$ in the dotted box, are notably lower than other $d$'s. 

\begin{figure*}
	\includegraphics[width=1.0\textwidth]{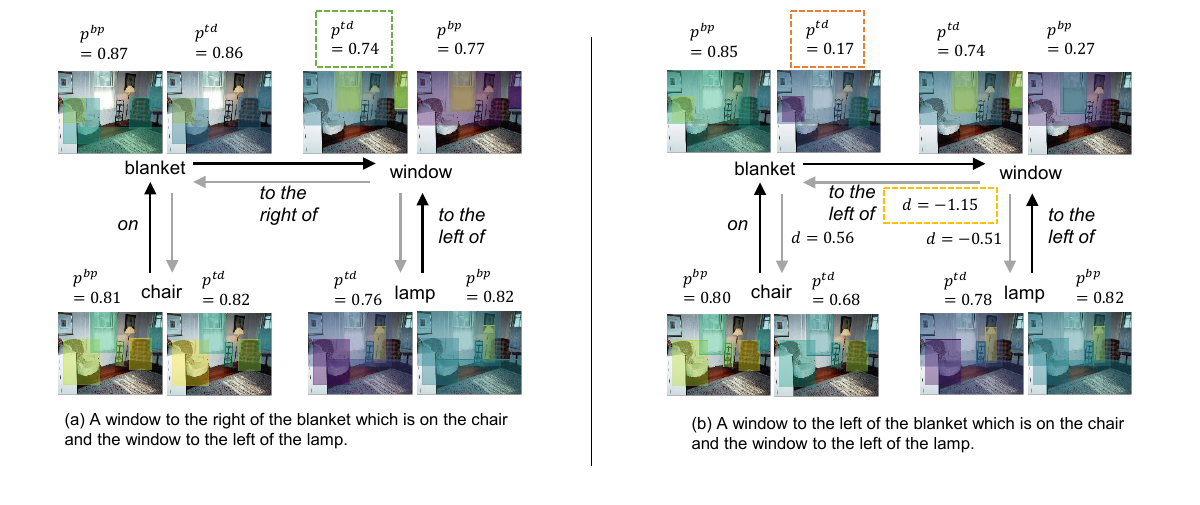}
	\caption{
		Two additional visualization showcases of RCRN. (a) is a matched case and (b) is the corresponding mismatched case. Both of them show the matching propagation results. 
	}
	\label{fig:showcase1}
\end{figure*}

\begin{figure*}
	\includegraphics[width=1.0\textwidth]{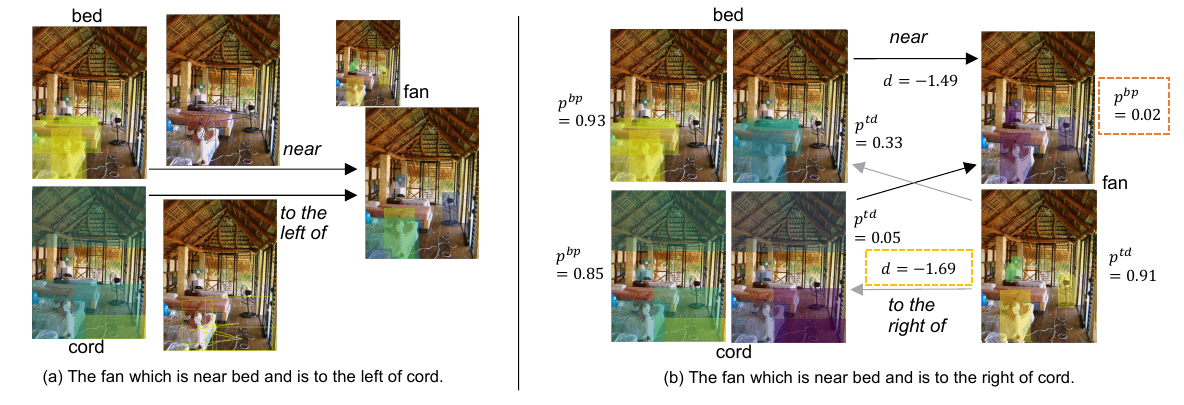}
	\caption{
		Another two cases of RCRN. (a) is the grounding process in a matched case and (b) is the matching propagation result on the corresponding mismatched case. 
	}
	\label{fig:showcase2}
\end{figure*}

Fig.~\ref{fig:showcase1} and Fig.~\ref{fig:showcase2} show some additional visualizations on RCRN. We highlight some inspiring evidences here. In Fig.~\ref{fig:showcase1}, the tiny relation mismatch causes notably drop on the matching probabilities of the attached nodes. In Fig.~\ref{fig:showcase2}, the ambiguity of ``fan" is significantly reduced by the message propagation, mainly relied on the message from the child branch ``to the left of cord". 

			
Tab.\textcolor{red}{\ref{tb:itp}} demonstrates some interpretable intermediate diagnosis results of RCRN on a subset of the validation set, with ground-truth box proposals. We analyze from the aspect of phrase-level visual grounding recall and statistics on relation correspondence scores. 
To obtain the true correspondence between the language scene graph and the bounding boxes,
we search the parsed language scene graph in the original image scene graph in GQA dataset~\cite{hudson2019gqa} with a noisy graph matching algorithm.  
See Sec.~\ref{sec:alg} for the detail of the algorithm.

The results verify that reasonable local correspondences are built, since the mean of initial recall (i.e. inference without MP) on entity phrase grounding is much higher than the random guess. As for relation correspondences, the average values on the ground-truth vision-language relation correspondences are far above the average mean value on all the pairs and very close to the average maximum. Moreover, message propagation notably helps refine the correspondence maps, especially on the referent entities.

\begin{table}[t]
	
	\centering
	\renewcommand\arraystretch{1.3} 
	\resizebox{0.48\textwidth}{!}{
		\begin{tabular}{c|ccc|ccc|ccc}
			\toprule[1pt]
			\textbf{Inference}     &   \multicolumn{3}{|c|}{\textbf{E.G.R.}}& \multicolumn{3}{|c|}{\textbf{R.G.R.}}& \multicolumn{3}{|c}{\textbf{R.C.S.}} \\
			\textbf{Method}   &\textbf{R@1} &\textbf{R@3}  &\textbf{R@5}  &\textbf{R@1} &\textbf{R@3}  &\textbf{R@5}  &\textbf{GT} &\textbf{Mean}  &\textbf{Max}\\ 
			
			\hline\hline
			By chance           &6.37&18.25&28.78&5.74&16.94&27.22& -& - & - \\        
			\hline
			w/o MP            &50.50	&80.59	&89.79          &38.88  &75.84  &88.11      &0.61  &0.17  &0.69\\                                
			w/ MP             &60.82 &85.58 &92.16 &69.46 &90.92 &95.59 &0.61  &0.17  &0.69\\
			\bottomrule[1pt]
		\end{tabular}
	}
	
	\caption{\small{Intermediate diagnosis results of RCRN. E.G.R. means Entity Grounding Recall. R.G.R. means Referent Grounding Recall. R.C.S. means Relation Correspondence Scores.
	}}
	\label{tb:itp}
\end{table}

\subsubsection{Generation of GT Correspondence}
\label{sec:alg}
We propose an algorithm mentioned above to find the ground-truth scene graph of the expression from the ground-truth image scene graph. 
The basic idea is that as the expressions from Ref-Reasoning are generated from some certain forms of the subgraphs of the image scene graphs in GQA \cite{hudson2019gqa}, the original subgraphs must exist in the complete scene graphs. We can achieve the golden subgraph by searching for the subgraph that is most close to the parsed language graph. 
The main work flow of the algorithm is shown as Alg. \ref{alg}.

Here we still suppose the language scene graph has a tree structure. The function $\operatorname{searchSuccessors}$ expands the current subgraph $g$ from their leaves, and only accepts the successors that matches the corresponding part in the parsed graph $(\V, \E)$, which is easy to verify by traversal. Note that the predicted subgraph $\hat{\G}$ may not exist, or not be unique, since the parsed language graph may have error. We only use the cases with a unique result for the subsequent procedures to find golden correspondences. Finally, we obtain the correspondences from the parsed phrases to visual components by those subgraphs, using the ground-truth correspondences annotated in the complete image scene graph.

\begin{algorithm}[t]
	\caption{The graph matching algorithm.}
	\label{alg}
	\small{
		\begin{algorithmic}[1]
			\Require
			The ground-truth coordinate of the referent $b$;
			The parsed entity phrases $\V$;
			The parsed relation phrases $\E$;
			The ground-truth scene graph objects $\V^\ast$ with phrase annotations;
			The ground-truth scene graph relations $\E^\ast$ with phrase annotations;
			\Ensure	  
			The predicted language scene graph $\hat{\G}$;
			
			\State stack $\leftarrow$ [$(\left\lbrace v \right\rbrace , \left\lbrace \right\rbrace )$ for $v$ in $\V^\ast$ if $\operatorname{coordinateOf}(v) == b$]
			
			\While {len(stack) $> 0$}
			\State $g \leftarrow$ stack.pop()
			
			\If {$g == (\V, \E)$}
			\State $\hat{\G} \leftarrow g$ 
			\State \Return $\hat{\G}$
			\Else
			\State stack.extend($\operatorname{searchSuccessors}$($g$, $\V$, $\E$, $\V^\ast$, $\E^\ast$))
			\EndIf
			\EndWhile
			%
			%
			%
			%
			%
			
	\end{algorithmic}}
\end{algorithm}
\newpage

\newpage

\end{document}